\documentclass{article}
\usepackage[margin=1in]{geometry}
\usepackage{amsmath,amssymb,amsthm}
\usepackage{graphicx}
\usepackage{tikz}
\usetikzlibrary{arrows.meta,positioning,calc,fit,backgrounds,shapes.geometric}
\usepackage{booktabs}
\usepackage[numbers]{natbib}
\usepackage{hyperref}
\usepackage{cleveref}
\usepackage{tabularx}
\usepackage{array}
\usepackage[table,dvipsnames]{xcolor}
\usepackage{placeins}

\newtheorem{theorem}{Theorem}

\hypersetup{breaklinks=true,colorlinks=true}

\begin{document}

\title{Naju: A Native Discrete State-Space Model with Independent Retention and Writing for Long-Sequence Memory}

\author{
Hyuk Lim and Seunghyun Yoon \\
Korea Institute of Energy Technology (KENTECH) \\
\texttt{hlim@kentech.ac.kr}, \texttt{syoon@kentech.ac.kr}
}

\maketitle

\begin{abstract}
Long-sequence memory tracking places two opposing demands on a recurrent state:
near-lossless \emph{retention} of stored bindings over long horizons, and active
\emph{overwriting} of stale ones. In our diagnostic suite, the strongest
efficient baselines tend to solve only one side well. We show that a
\emph{native discrete} state-space model (SSM) can do both at once. Continuous-time-parameterized SSMs such
as Mamba obtain their discrete recurrence by zero-order-hold discretization of
a continuous-time system; we argue that this detour is unnecessary for memory
tracking and parameterize the discrete transition directly. \textbf{Naju}
(Native Adaptive Junction Unit) factorizes the recurrent update, schematically
$x_n = f_n\odot x_{n-1} + i_n\odot(B_n u_n)$, into an explicit discrete pole
(a learned forget gate $f_n$), an independent write gain $i_n$, and
input-dependent write/read maps. Since the sigmoid pole satisfies
$0<f_n<1$, each frozen local coordinate is Schur-stable by construction, and
the full time-varying recurrence satisfies a fading-memory/BIBO bound under
uniform boundedness assumptions, with no stability regularizer. We formalize the key structural limitation of coupled designs: any
non-expansive complementary single-gate recurrence ties the effective
retention $r$ and write gain $w$ through $|r|+w\le 1$, so near-complete
retention forces weak writing; decoupling $f_n$ from $i_n$ removes this
constraint. Empirically, Naju is the only evaluated model that is strong on both axes at $4\times$ the training length
(retention $0.99$ $\wedge$ overwrite $0.89$), while the strongest baselines
specialize in one side. Beyond the diagnostic suite, at a matched $1.2$B-token budget on
WikiText-103, Naju achieves the lowest perplexity at
$d_{\mathrm{model}}=256$ across context lengths from $1{,}024$ to $4{,}096$,
and remains competitive with the Transformer while outperforming both Mamba
generations as the model width scales to $1{,}024$. It also achieves the highest average
among the fully evaluated models in our budget-matched Long Range Arena
comparison and, at
$d_{\mathrm{model}}=256$, outperforms both Mamba generations under matched
recurrent-state budgets on multi-query associative recall, while retaining
the linear-time, linear-memory scaling of an SSM. Together, these
results support a simple principle for long-sequence memory in discrete state
spaces: give retention and writing their own gates---one to hold and an
independent one to write---so that a single fixed-size recurrent state can both
preserve and update its bindings over long sequences.
\end{abstract}

\section{Introduction}

Transformers dominate modern sequence modeling, but full self-attention scales quadratically with sequence length, limiting its use in long-context settings. State space models (SSMs), most notably Mamba \cite{gu2023mamba}, provide an alternative with linear-time sequence mixing through input-dependent state transitions implemented by hardware-aware parallel scans.
Most SSMs begin from a continuous-time system,
\begin{equation}
\dot{h}(\tau)=Ah(\tau)+Bu(\tau),
\qquad
y(\tau)=Ch(\tau),
\end{equation}
and obtain a discrete recurrence through zero-order-hold (ZOH) discretization. Given a step size $\Delta_n$,
\begin{equation}
\bar{A}_n=\exp(\Delta_n A),
\qquad
\bar{B}_n=(\Delta_n A)^{-1}
\left(\exp(\Delta_n A)-I\right)\Delta_n B.
\end{equation}
Selective SSMs make $\Delta_n$, $B_n$, and $C_n$ functions of the input. This continuous-time construction provides a principled route to stable and structured dynamics. For the token, event, and state-update sequences ultimately processed by the model, however, the operational object is the discrete recurrence
\begin{equation}
x_n=\bar{A}_n x_{n-1}+\bar{B}_n u_n.
\end{equation}
This raises a basic design question: for discrete sequence memory, should the recurrent transition itself be parameterized directly?

We study this question in the setting of \emph{long-sequence memory tracking}: storing many bindings and later recalling, updating, or overwriting them after long intervals. This setting arises in practical workloads rather than only in synthetic recall tasks. Long-session dialogue agents must preserve user facts and instructions while processing many intervening turns; entity-consistent generation and summarization require a binding introduced once to survive until it is queried or revised; repository-level code models encounter identifiers defined once and referenced far later; and streaming analysis of system logs, security events, or industrial telemetry must maintain the latest state of an entity amid large numbers of unrelated updates. These applications share two demands under a fixed state budget: important bindings must be retained with little attenuation, yet stale bindings must be overwritten decisively when their values change.

The first demand is governed by the recurrent \emph{pole}---the multiplicative factor
that controls the temporal decay of the previous state. For a scalar frozen recurrence with retention $f$, the contribution of a stored value after $L$ steps is proportional to $f^L$. Retaining at least a fraction $\rho$ therefore requires
\begin{equation}
f \geq \rho^{1/L},
\end{equation}
or equivalently $1-f=O(1/L)$ for a nonvanishing retained contribution over increasing horizons. Long-range memory consequently requires an effective pole extremely close to one.
A continuous-time SSM can enter this regime because
$\bar{A}_n=\exp(\Delta_n A)\rightarrow I$ as $\Delta_n\rightarrow0$. Under ZOH, however, the same step size also affects the discretized input map:
\begin{equation}
\bar{A}_n=I+\Delta_n A+O(\Delta_n^2),
\qquad
\bar{B}_n=\Delta_n B+O(\Delta_n^2).
\end{equation}
Thus, in the small-step regime, moving the transition toward identity also reduces the immediate write scale unless other parameters compensate for it. The issue is not that a ZOH-based model is incapable of retaining or writing, but that retention and input injection are influenced by the same discretization variable. This creates an optimization geometry in which the two functions are not directly controlled by independent parameters.

A second form of coupling arises when explicit temporal control is added on
top of an SSM transition through a complementary gate,
\begin{equation}
x_n
=
(1-g_n)\odot\left(\bar{A}_n x_{n-1}\right)
+
g_n\odot\left(\bar{B}_n u_n\right),
\qquad
g_n=\sigma(\cdot).
\end{equation}
Here, the effective recurrent transition is
$(1-g_n)\odot\bar{A}_n$. Because $1-g_n<1$ for finite logits, the gate
introduces an additional subunit factor into the retention path. More
importantly, the same scalar or channel-wise variable controls two opposing
quantities: increasing $g_n$ strengthens writing but suppresses retention
through $1-g_n$, whereas decreasing $g_n$ strengthens retention but attenuates
the current write. Such a gate can alternate between retaining and updating,
but it cannot choose the retention and write gains independently at the same
step. 

We therefore propose \textbf{Naju}, a native-discrete selective state-space layer in which the recurrent pole and write gain are parameterized separately:
\begin{equation}
x_n
=
f_n\odot x_{n-1}
+
i_n\odot\left(B_n h_n\right),
\label{eq:naju-intro}
\end{equation}
where $f_n=\sigma(\cdot)$ is the forget gate and $i_n=\sigma(\cdot)$ is an independent input gate. There is no continuous-time generator, step size $\Delta_n$, matrix exponential, or ZOH map: $f_n$ is itself the discrete recurrent pole. The model can therefore represent a near-identity transition with $f_n\rightarrow1$ without forcing the write gain $i_n$ toward zero. Conversely, it can overwrite stale state by setting $f_n\rightarrow0$ and $i_n\rightarrow1$. More generally, it can preserve existing state while adding new information by allowing both gates to be large.

Naju adopts the retain/write separation associated with the LSTM memory cell, but it does not simply insert an LSTM cell into an SSM block. In an LSTM, the gates update a cell state whose contents are exposed through a nonlinear hidden-state transformation. In Naju, the forget gate is the pole of a diagonal selective SSM state, $i_n B_n u_n$ is an input-dependent state write, and $C_n^{\top}x_n$ is an input-dependent state readout. The retain/write junction is therefore realized as a native-discrete, selective, and associative-scan-compatible state-space recurrence.

Direct discrete-time modeling alone is not unique to Naju. Gated linear recurrences such as HGRN, RetNet, the LRU family, and Griffin's RG-LRU also avoid a continuous-time discretization step. Their mechanisms, however, differ from the combination studied here. Complementary-gated recurrences directly tie the write coefficient to the retention gate; LRU-style models use a retention-dependent normalization such as $\sqrt{1-\lambda_n^2}$; and fixed- or structured-decay models do not provide an independent pair of input-dependent retain and write gates acting on a selective state. Naju combines three properties that are not jointly present in these models: the forget gate is the discrete pole itself, the write gain is independently learned, and input-dependent $B_n$ and $C_n$ selectively write to and read from the state.

This distinction is central to long-sequence memory tracking because retention and overwriting impose different requirements. Holding a binding across a long interval requires $f_n$ to remain close to one over the relevant steps. Replacing a stale binding requires the model to suppress the old state and inject the new value, corresponding to $f_n\rightarrow0$ and $i_n\rightarrow1$. A complementary gate can move between these modes but imposes a fixed relationship between them. Naju instead learns the two decisions separately and can allocate different channels and time steps to preservation, accumulation, or replacement.

\begin{figure}[t]
\centering
\includegraphics[width=0.55\linewidth]{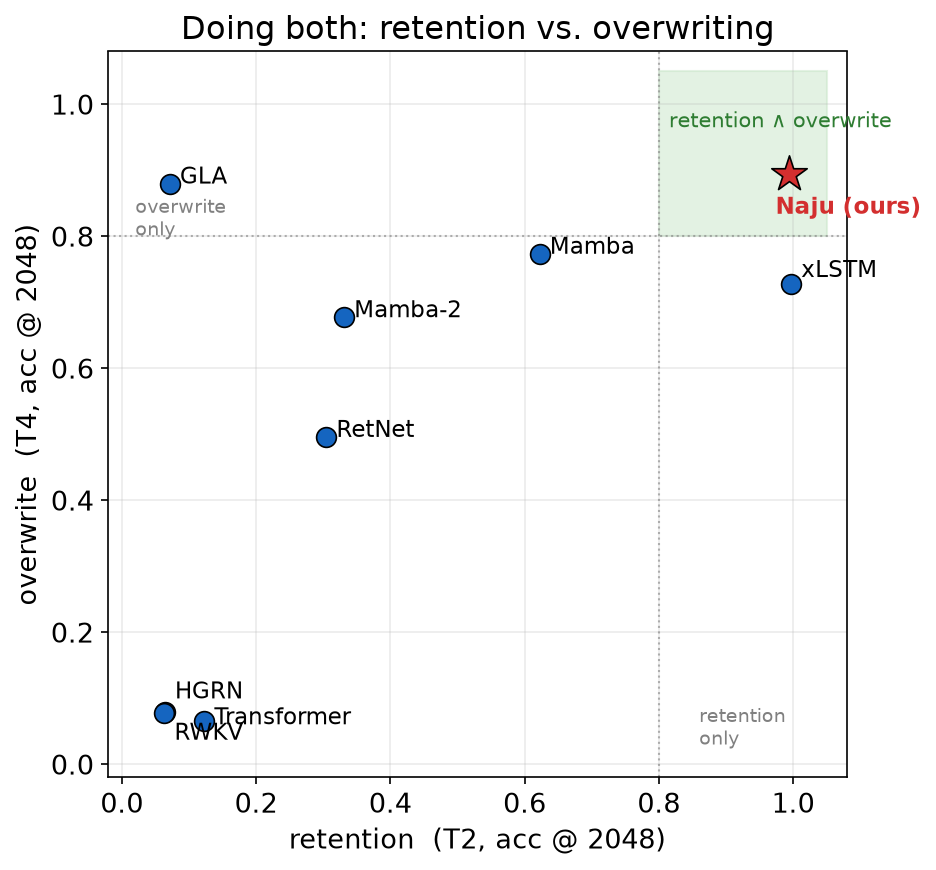}
\caption{\textbf{Retention and overwriting at length 2048.}
Extrapolation accuracy on retention (T2, scattered key--value recall; $x$-axis) and overwriting (T4, hard state tracking; $y$-axis) under the shared training protocol and the stated model configurations. Among the tested baselines, xLSTM provides strong retention but weaker overwriting, whereas GLA provides strong overwriting but weak long-distance retention. Mamba and Mamba-2 do not reach either extreme simultaneously. Naju occupies the top-right region by independently controlling its forget and input gates.}
\label{fig:frontier}
\end{figure}

Figure~\ref{fig:frontier} illustrates the resulting empirical distinction. At length $2048$, the strongest tested baselines tend to specialize in one side of the problem: xLSTM attains retention accuracy of $1.00$ but overwriting accuracy of $0.73$, whereas GLA attains overwriting accuracy of $0.88$ but retention accuracy of only $0.07$. Mamba obtains $0.62/0.77$ on the two axes. Naju reaches $0.99/0.89$, making it the only tested model in the high-retention, high-overwriting region. We use this ability to perform both functions within one recurrent state as the central empirical criterion of the paper.

Our contributions are as follows:
\begin{enumerate}
\item We characterize two sources of retain--write coupling in efficient recurrent sequence models. In the small-step regime of ZOH-discretized SSMs, the step size that moves the transition toward identity also scales the discretized input map. In complementary temporal gating, the same variable strengthens retention while suppressing writing. We formalize how these parameterizations affect the long-range memory kernel and associative recall.

\item We introduce Naju, a native-discrete selective state-space layer whose forget gate is the recurrent pole and whose input gate independently controls writing. Naju combines independent retain/write gains with input-dependent state-write and state-read maps $B_n$ and $C_n$, generated with short causal local context, while preserving a diagonal recurrence compatible with associative parallel scans.

\item We analyze Naju as a discrete dynamical system. Under frozen gates, each state coordinate is a first-order discrete IIR filter whose pole is exactly the corresponding forget-gate value. For the time-varying recurrence, we establish a fading-memory and BIBO-stability bound under uniformly bounded gate logits, input drives, and readout parameters, without requiring an additional stability regularizer. We also relate the resulting recurrence to the gated-RNN limit of selective SSMs.

\item We show empirically that, among the tested models, Naju is uniquely strong at both long-range retention and active overwriting. It achieves nearly distance-uniform associative recall with approximately $1.1$M parameters and occupies the top-right retention--overwriting region at length $2048$. The advantage also transfers beyond diagnostic tasks: under a shared six-layer, $d_{\mathrm{model}}=256$ language-modeling protocol on WikiText-103, Naju obtains a test perplexity of $26.20\pm0.17$, compared with $28.31\pm0.17$ for Mamba-2, while retaining linear-time recurrent execution.

\end{enumerate}

\section{Related Work}
\label{sec:related_work}

\subsection{Structured and Selective State Space Models}

State space models (SSMs) represent a sequence through a recurrent state update
and a state-dependent readout. Modern structured SSMs make this formulation
practical for long sequences by imposing diagonal or otherwise structured
transition operators that admit efficient convolutional or parallel-scan
implementations. H3~\cite{fu2023h3} adapts structured SSMs to language modeling
and identifies associative recall as a central limitation of time-invariant
SSMs. It supplements the recurrent state with a shift operation so that nearby
tokens can interact before being written into memory.

Mamba~\cite{gu2023mamba} introduces input-dependent selectivity into the SSM
recurrence and implements it with a hardware-aware scan. Its model is
parameterized through continuous-time dynamics and discretized at each token,
with the input-dependent step size $\Delta_n$ jointly affecting the discrete
transition $\bar{A}_n$ and input map $\bar{B}_n$, while $B_n$ and $C_n$ provide
additional input-dependent write and read selectivity. Mamba-2 and structured
state-space duality~\cite{dao2024mamba2} further connect selective SSMs with
structured masked attention and reorganize the computation for improved
hardware efficiency.

Naju retains the selective write/read structure of this family but
parameterizes the recurrent dynamics directly in discrete time. Its forget
gate $f_n$ is the discrete recurrent pole rather than a factor derived from
$\exp(\Delta_n A)$, and its independently parameterized input gate $i_n$
controls the magnitude of the selective write. Naju therefore changes the
parameterization of the recurrent core rather than replacing selective state
maps or scan-based execution.

\subsection{Gated Recurrent and Linear-Time Sequence Models}

The separation of memory retention and input writing originates in gated
recurrent architectures. An LSTM updates its memory cell according to
\begin{equation}
\begin{aligned}
c_n
&=
f_n \odot c_{n-1}
+
i_n \odot \widetilde{c}_n, \\
h_n
&=
o_n \odot \tanh(c_n),
\end{aligned}
\end{equation}
where the forget and input gates separately control retention and writing.
This separation improves memory control compared with a vanilla RNN, whose
retention and input processing are entangled in a dense nonlinear hidden-state
transition.

Recent linear-time architectures revisit gated recurrence using simpler or
more structured state updates. RetNet~\cite{sun2023retnet} uses multiplicative
retention to obtain parallel, recurrent, and chunkwise computation forms.
RWKV~\cite{peng2023rwkv,peng2024rwkv6} combines recurrent weighted
key--value aggregation with token-shift and channel-mixing mechanisms.
GLA~\cite{yang2024gla} introduces data-dependent multiplicative gates into a
linear-attention state, whereas HGRN~\cite{qin2024hgrn} uses a gated linear
recurrence with hierarchically constrained retention. These models demonstrate
that explicit control of recurrent decay is important for efficient
long-context processing, although their state organizations and write
mechanisms differ.

The LRU~\cite{orvieto2023lru} shows that carefully parameterized stable
diagonal linear recurrences can perform strongly on long-range benchmarks.
Its transition is linear and time invariant, and a retention-dependent
normalization is used to balance state propagation and input injection.
Griffin's RG-LRU~\cite{de2024griffin}, subsequently used in
RecurrentGemma~\cite{botev2024recurrentgemma}, makes both the recurrence
coefficient and an input gate data dependent. A representative scalar form
of its update is
\begin{equation}
h_n
=
\lambda_n \odot h_{n-1}
+
\sqrt{1-\lambda_n^2}
\odot
\left(i_n \odot u_n\right).
\end{equation}
The input gate $i_n$ provides data-dependent write control, but the admissible
write amplitude remains modulated by the retention-dependent envelope
$\sqrt{1-\lambda_n^2}$. Consequently, as $\lambda_n$ approaches one, the
normalized input contribution approaches zero regardless of the value of
$i_n$. This is a deliberate normalization and stabilization choice rather than
an absence of input gating, but it differs from Naju's independently
parameterized retention and write gains.

xLSTM~\cite{beck2024xlstm} is closest to Naju in its explicit use of distinct
forget and input gates. Its sLSTM and mLSTM variants extend the LSTM framework
with exponential gating, normalization states, and scalar or matrix-valued
memories. These mechanisms enable expressive recurrent memory but introduce a
different state structure and stabilization procedure. Naju instead uses a
sigmoid-gated diagonal recurrence whose local pole lies in $(0,1)$ for finite
logits and whose state update remains affine and compatible with an
associative scan.

Naju can be related to the LSTM principle without being identified with a
conventional LSTM cell. Its recurrent core is
\begin{equation}
\begin{aligned}
x_n
&=
f_n \odot x_{n-1}
+
i_n \odot \left(B_n h_n\right), \\
y_n
&=
\alpha C_n^{\top}x_n
+
D h_n .
\end{aligned}
\end{equation}
Here, $x_n$ is a diagonal SSM state, $B_n$ and $C_n$ are input-dependent
selective write/read maps, and $\alpha$ is a constant gain on the state-readout
path. Thus, the LSTM and Naju recurrences share the principle of separately
parameterized retention and writing, but they differ in the memory carrier,
readout mechanism, and computation graph. Under frozen parameters, each Naju
state coordinate is a first-order discrete IIR filter, and the full
time-varying update remains an affine diagonal scan.

The output modulation in Naju is provided by a GLU/SwiGLU-style
branch~\cite{dauphin2017glu,shazeer2020glu}, as commonly used in modern
sequence blocks. This branch plays a role analogous to output gating without
changing the affine recurrent state update. Naju therefore combines
forget-, input-, and output-like control while preserving a selective
SSM-style memory carrier and an associative-scan-compatible recurrence.

\begin{table*}[t]
\centering
\small
\setlength{\tabcolsep}{4.0pt}
\renewcommand{\arraystretch}{1.10}
\begin{tabularx}{\textwidth}{
    >{\raggedright\arraybackslash}p{0.12\textwidth}
    >{\raggedright\arraybackslash}p{0.16\textwidth}
    >{\raggedright\arraybackslash}p{0.18\textwidth}
    >{\raggedright\arraybackslash}p{0.21\textwidth}
    >{\raggedright\arraybackslash}X
}
\toprule
\textbf{Model family}
&
\textbf{Memory carrier}
&
\textbf{Retention}
&
\textbf{Input writing}
&
\textbf{Key distinction from Naju}
\\
\midrule

LSTM
&
Nonlinear cell state $c_n$
&
Independent forget gate $f_n$
&
Independent input gate applied to a nonlinear candidate
&
No selective SSM write/read maps; nonlinear hidden-state readout
\\

Mamba
&
Selective SSM state
&
$\bar{A}_n=\exp(\Delta_n A)$ after discretization
&
Discretized selective map $\bar{B}_n u_n$
&
$\Delta_n$ participates in both the discrete transition and input map
\\

HGRN and related complementary-gated recurrences
&
Vector or structured recurrent state
&
Data-dependent retention gate
&
Write coefficient coupled to the retention gate, often through a
complementary term
&
Retention and writing are not independently parameterized
\\

LRU / RG-LRU
&
Diagonal linear recurrent state
&
Stable diagonal coefficient, optionally data dependent
&
Input gate with retention-dependent normalization
&
The write amplitude has a retention-dependent envelope
\\

xLSTM
&
Scalar or matrix-valued LSTM memory
&
Independent exponential forget gate
&
Independent exponential input gate with normalization states
&
Different memory structure, normalization, and nonlinear output mechanism
\\

Naju
&
Diagonal selective SSM state
&
Forget gate $f_n$ used directly as the discrete pole
&
Independent gate $i_n$ applied to the selective write $B_n u_n$
&
Native-discrete pole, independent retain/write gains, and selective
$B_n$ and $C_n$
\\

\bottomrule
\end{tabularx}
\caption{
Structural positioning of Naju relative to representative recurrent sequence
models. The table distinguishes independently parameterized gates from write
mechanisms whose scale remains coupled to retention through a complementary
coefficient, discretization variable, or normalization envelope. These
categories describe parameterization rather than absolute expressive
limitations: sufficiently flexible models may compensate through other
parameters, layers, or channels.
}
\label{tab:rnn_lstm_mamba_naju}
\end{table*}

\subsection{Hybrid and Long-Context Architectures}

Hybrid architectures combine recurrent or state-space layers with attention to
balance efficient sequence processing and content-based retrieval.
Griffin/Hawk~\cite{de2024griffin} combines RG-LRU blocks with local attention,
while Jamba~\cite{lieber2024jamba} interleaves Transformer attention and Mamba
layers. These architectures show that recurrent sequence mixers can be
effective components of larger systems when complemented by local or sparse
attention.

Naju addresses a complementary question. Rather than adding attention around
an existing recurrent layer, it modifies the recurrent parameterization
itself. The objective is to determine whether a native-discrete selective
recurrence with separately parameterized retain and write gains can provide
both long-range preservation and decisive state replacement. We therefore
evaluate the recurrent core directly on associative recall, state tracking,
length extrapolation, language modeling, and throughput, without attributing
its memory behavior to an interleaved attention mechanism.

\section{Method: Naju, a Native Discrete SSM}
\label{sec:method}

We index sequence positions by $n=1,2,\ldots,L$, where $L$ is the
sequence length, and layers by $l=1,2,\ldots,M$. Unless otherwise stated,
the equations below describe a single layer, and the layer superscript is
included only when needed. Naju maintains a diagonal recurrent state for each
expanded channel.

\subsection{Discrete SSM Background}
\label{sec:ssm_background}

For a single channel, a diagonal state-space layer maintains a recurrent state
$x_n\in\mathbb{R}^{d_{\text{state}}}$ and applies
\begin{equation}
x_n
=
\bar{A}_n x_{n-1}
+
\bar{B}_n u_n,
\qquad
y_n
=
C_n^{\top}x_n .
\label{eq:ssm_background}
\end{equation}
Here, $\bar{A}_n$ is the discrete transition, $\bar{B}_n$ maps the current
input into the state, and $C_n$ reads from the state.

In continuous-time-parameterized SSMs such as Mamba, the discrete recurrence
is obtained by first defining a continuous-time state-space system and then
applying a discretization rule. Under zero-order-hold (ZOH) discretization,
the discrete transition is
\begin{equation}
\bar{A}_n
=
\exp\left(\Delta_n A\right),
\label{eq:zoh_A}
\end{equation}
and the corresponding input map can be written as
\begin{equation}
\bar{B}_n
=
\left(
\int_{0}^{\Delta_n}
\exp\left((\Delta_n-s)A\right)
\,ds
\right)
B_n .
\label{eq:zoh_B}
\end{equation}
Mamba introduces selectivity by making $\Delta_n$, $B_n$, and $C_n$
input dependent. Consequently, the discrete transition and input map vary
with the token while retaining the continuous-time-to-discrete construction.

Naju takes a different route. Rather than defining a continuous-time operator
and discretizing it at every step, Naju parameterizes the recurrence directly
in discrete time. Its memory retention is controlled by a learned forget gate
$f_n$, which acts directly as the discrete pole, while state writing is
controlled separately by an input gate $i_n$. Thus, Naju uses neither a
learned step size $\Delta_n$ nor a ZOH discretization map.

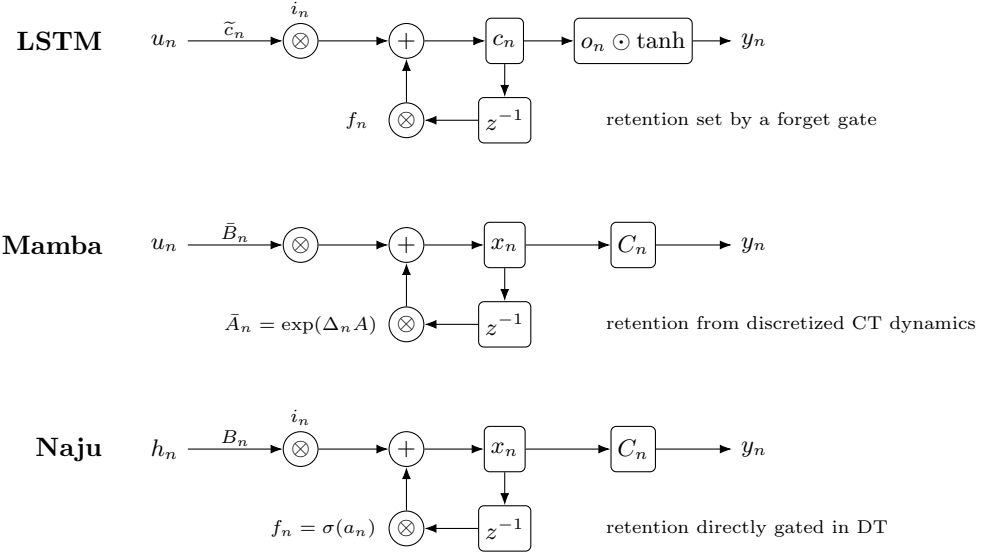
\begin{figure}[t]
\centering
\begin{tikzpicture}[
    font=\small,
    op/.style={
        circle,
        draw,
        inner sep=0pt,
        minimum size=4.6mm
    },
    box/.style={
        draw,
        rounded corners=2pt,
        minimum height=6mm,
        inner sep=2.5pt
    },
    sig/.style={
        -{Latex[length=1.6mm]}
    },
    gl/.style={
        font=\scriptsize,
        inner sep=1pt
    }
]

\begin{scope}[yshift=0cm]
    \node[anchor=east,font=\bfseries] at (-0.7,0) {LSTM};

    \node (in) at (0,0) {$u_n$};
    \node[op] (mi) at (1.8,0) {$\otimes$};
    \node[op] (sum) at (3.2,0) {$+$};
    \node[box] (st) at (4.5,0) {$c_n$};
    \node[box] (rd) at (6.2,0) {$o_n\odot\tanh$};
    \node (out) at (7.8,0) {$y_n$};

    \node[op] (mf) at (3.2,-1.05) {$\otimes$};
    \node[box] (dly) at (4.5,-1.05) {$z^{-1}$};

    \draw[sig]
        (in)
        --
        node[gl,above] {$\widetilde{c}_n$}
        (mi);

    \draw[sig] (mi) -- (sum);
    \draw[sig] (sum) -- (st);
    \draw[sig] (st) -- (rd);
    \draw[sig] (rd) -- (out);
    \draw[sig] (st) -- (dly);
    \draw[sig] (dly) -- (mf);
    \draw[sig] (mf) -- (sum);

    \node[gl] at (1.8,0.42) {$i_n$};
    \node[gl] at (2.55,-1.05) {$f_n$};

    \node[gl,anchor=west] at (5.8,-1.05)
        {retention set by a forget gate};
\end{scope}

\begin{scope}[yshift=-2.7cm]
    \node[anchor=east,font=\bfseries] at (-0.7,0) {Mamba};

    \node (in) at (0,0) {$u_n$};
    \node[op] (mi) at (1.8,0) {$\otimes$};
    \node[op] (sum) at (3.2,0) {$+$};
    \node[box] (st) at (4.5,0) {$x_n$};
    \node[box] (rd) at (6.2,0) {$C_n$};
    \node (out) at (7.8,0) {$y_n$};

    \node[op] (mf) at (3.2,-1.05) {$\otimes$};
    \node[box] (dly) at (4.5,-1.05) {$z^{-1}$};

    \draw[sig]
        (in)
        --
        node[gl,above] {$\bar{B}_n$}
        (mi);

    \draw[sig] (mi) -- (sum);
    \draw[sig] (sum) -- (st);
    \draw[sig] (st) -- (rd);
    \draw[sig] (rd) -- (out);
    \draw[sig] (st) -- (dly);
    \draw[sig] (dly) -- (mf);
    \draw[sig] (mf) -- (sum);

    \node[gl] at (1.8,-1.05)
        {$\bar{A}_n=\exp(\Delta_n A)$};

    \node[gl,anchor=west] at (5.8,-1.05)
        {retention from discretized CT dynamics};
\end{scope}

\begin{scope}[yshift=-5.4cm]
    \node[anchor=east,font=\bfseries] at (-0.7,0) {Naju};

    \node (in) at (0,0) {$h_n$};
    \node[op] (mi) at (1.8,0) {$\otimes$};
    \node[op] (sum) at (3.2,0) {$+$};
    \node[box] (st) at (4.5,0) {$x_n$};
    \node[box] (rd) at (6.2,0) {$C_n$};
    \node (out) at (7.8,0) {$y_n$};

    \node[op] (mf) at (3.2,-1.05) {$\otimes$};
    \node[box] (dly) at (4.5,-1.05) {$z^{-1}$};

    \draw[sig]
        (in)
        --
        node[gl,above] {$B_n$}
        (mi);

    \draw[sig] (mi) -- (sum);
    \draw[sig] (sum) -- (st);
    \draw[sig] (st) -- (rd);
    \draw[sig] (rd) -- (out);
    \draw[sig] (st) -- (dly);
    \draw[sig] (dly) -- (mf);
    \draw[sig] (mf) -- (sum);

    \node[gl] at (1.8,0.42) {$i_n$};
    \node[gl] at (2.1,-1.05) {$f_n=\sigma(a_n)$};

    \node[gl,anchor=west] at (5.8,-1.05)
        {retention directly gated in DT};
\end{scope}

\end{tikzpicture}
\caption{
Comparison of the recurrent junctions in LSTM, Mamba, and Naju.
Each recurrence contains an additive node combining a retained state with a
current write. LSTM controls cell-state retention with a forget gate. Mamba
obtains its discrete transition from a continuous-time-parameterized SSM,
for example, $\bar{A}_n=\exp(\Delta_n A)$. Naju instead uses the learned
forget gate $f_n$ directly as the discrete pole and controls writing with a
separate input gate $i_n$. The figure compares only the recurrent updates:
readout gains, feedthrough paths, and outer residual connections are omitted.
For Naju, $h_n$ denotes the convolved content drive defined in
Eq.~\eqref{eq:naju_content_branch}.
}
\label{fig:compare}
\end{figure}

\subsection{Native Discrete Recurrence and Readout}
\label{sec:najudss}
\label{sec:readout}

The name \textbf{Naju} stands for \emph{Native Adaptive Junction Unit}.
It is \emph{native discrete} because its recurrence is parameterized directly
in discrete time, without introducing a continuous-time transition operator or
a zero-order-hold discretization step. It is \emph{adaptive} because its
retention and write strengths vary with the input. The term \emph{junction}
refers to the additive point at which the retained state and newly written
content meet.

Naju is defined by two central equations. For layer $l$, the recurrent state
is updated as
\begin{equation}
x_n^{(l)}
=
f_n^{(l)}
\odot
x_{n-1}^{(l)}
+
i_n^{(l)}
\odot
\left(
h_n^{(l)}
\left(B_n^{(l)}\right)^{\top}
\right),
\label{eq:naju_state_update}
\end{equation}
where
\begin{equation}
f_n^{(l)}
=
\sigma\left(a_n^{(l)}\right),
\qquad
i_n^{(l)}
=
\sigma\left(b_n^{(l)}\right).
\label{eq:naju_gates}
\end{equation}
After the recurrent scan, the state is read out as
\begin{equation}
y_n^{(l)}
=
\alpha\,
x_n^{(l)}C_n^{(l)}
+
D^{(l)}
\odot
h_n^{(l)} .
\label{eq:naju_readout}
\end{equation}
Table~\ref{tab:naju_core_notation} summarizes only the quantities that appear
directly in Eqs.~\eqref{eq:naju_state_update}--\eqref{eq:naju_readout}. We use
$d_{\text{inner}}=E d_{\text{model}}$, with $E=2$ in the main experiments.

\begin{table}[t]
\centering
\footnotesize
\setlength{\tabcolsep}{3.5pt}
\renewcommand{\arraystretch}{1.08}
\begin{tabularx}{\columnwidth}{
    >{\raggedright\arraybackslash}p{0.15\columnwidth}
    >{\raggedright\arraybackslash}p{0.15\columnwidth}
    >{\raggedright\arraybackslash}X
}
\toprule
\textbf{Symbol}
&
\textbf{Shape}
&
\textbf{Meaning}
\\
\midrule

$h_n^{(l)}$
&
$\mathbb{R}^{d_{\text{inner}}}$
&
Convolved content drive supplied to the write and feedthrough paths.
\\

$x_n^{(l)}$
&
$\mathbb{R}^{d_{\text{inner}}\times d_{\text{state}}}$
&
Recurrent memory; each expanded channel maintains a
$d_{\text{state}}$-dimensional state.
\\

$f_n^{(l)},\,i_n^{(l)}$
&
$(0,1)^{d_{\text{inner}}}$
&
Independent retain and write gates, broadcast across the state dimension.
\\

$B_n^{(l)},\,C_n^{(l)}$
&
$\mathbb{R}^{d_{\text{state}}}$
&
Token-dependent state-space write and read directions.
\\

$\alpha$
&
$\mathbb{R}_{>0}$
&
Constant gain applied to the recurrent-memory readout.
\\

$D^{(l)}$
&
$\mathbb{R}^{d_{\text{inner}}}$
&
Learned token-independent feedthrough gain.
\\

$y_n^{(l)}$
&
$\mathbb{R}^{d_{\text{inner}}}$
&
Local output before the block output projection.
\\

\bottomrule
\end{tabularx}
\caption{Core notation for the Naju state update and readout.}
\label{tab:naju_core_notation}
\end{table}

Equation~\eqref{eq:naju_state_update} consists of a \emph{retain branch} and a
\emph{write branch}. The retain branch,
$f_n^{(l)}\odot x_{n-1}^{(l)}$, propagates the previous state. The forget
gate $f_n^{(l)}$ is therefore the token-dependent discrete pole of the
recurrence: values close to one preserve stored information, whereas values
close to zero suppress the previous state.
The write branch,
\begin{equation}
i_n^{(l)}
\odot
\left(
h_n^{(l)}
\left(B_n^{(l)}\right)^{\top}
\right),
\end{equation}
injects the current content into the state. The drive $h_n^{(l)}$ contains
the local content to be written, while $B_n^{(l)}$ determines the
state-space direction of the write. Their outer product satisfies
\begin{equation}
h_n^{(l)}
\left(B_n^{(l)}\right)^{\top}
\in
\mathbb{R}^{d_{\text{inner}}\times d_{\text{state}}},
\end{equation}
matching the shape of the recurrent state. The gate $i_n^{(l)}$ independently
controls the write magnitude for each expanded channel.
The independence of $f_n^{(l)}$ and $i_n^{(l)}$ permits distinct memory
operations. When $f_n^{(l)}$ is large and $i_n^{(l)}$ is small, the unit
approximately holds its state. When $f_n^{(l)}$ is small and
$i_n^{(l)}$ is large, it suppresses stale state and writes new content.
When both gates are large, it can add new evidence while preserving previously
stored information. Retention and writing are therefore separate decisions
rather than complementary outcomes of a single gate.

In Eq.~\eqref{eq:naju_readout}, the product
$C_n^{(l)}x_n^{(l)}$ denotes contraction over the
$d_{\text{state}}$ dimension:
\begin{equation}
x_n^{(l)}C_n^{(l)}
\in
\mathbb{R}^{d_{\text{inner}}}.
\end{equation}
The token-dependent vector $C_n^{(l)}$ determines how the current state is
queried, while $\alpha$ controls the scale of the resulting recurrent-memory
contribution.
The feedthrough term
\begin{equation}
D^{(l)}
\odot
h_n^{(l)}
\end{equation}
provides a direct route from the current content drive to the local output.
The parameter $D^{(l)}$ is learned but token independent and acts diagonally
across expanded channels. This path allows local information to be exposed
without first storing and retrieving it through the recurrent state.

Neither $\alpha$ nor $D^{(l)}$ enters
Eq.~\eqref{eq:naju_state_update}. They therefore do not alter the recurrent
pole or the memory dynamics. The pole is determined by $f_n^{(l)}$, whereas
$\alpha$ and $D^{(l)}$ control the relative scales of the memory-mediated and
direct output paths. In a local transfer-function interpretation,
$f_n^{(l)}$ determines the pole, while the direct feedthrough term contributes
to the zero structure.

\paragraph{Relation to LSTM.}
Equation~\eqref{eq:naju_state_update} shares the abstract retain/write pattern
of the LSTM memory update
\begin{equation}
c_n
=
f_n
\odot
c_{n-1}
+
i_n
\odot
\widetilde{c}_n,
\end{equation}
but its memory carrier and computation graph are different. Naju maintains a
diagonal selective SSM state rather than an LSTM cell exposed through a
nonlinear hidden-state transformation. Its write content is the selective
outer product $h_nB_n^{\top}$ rather than a dense nonlinear candidate cell,
and its state is queried through the token-dependent read vector $C_n$.
Naju should therefore be understood as an SSM-style realization of
retain/write decoupling rather than as a conventional LSTM cell.

\paragraph{Native-discrete SSM interpretation.}
The roles of the principal quantities are explicit: $f_n$ controls temporal
retention, $i_n$ controls write magnitude, $B_n$ determines the state-space
write direction, and $C_n$ determines the state-space read direction. The
scalar $\alpha$ and feedthrough gain $D$ affect the output scale but not the
state recurrence.
Compared with continuous-time-parameterized selective SSMs, the principal
structural difference is that retention is not mediated by
$\exp(\Delta_n A)$ or by another discretization map. Instead, the discrete
pole is parameterized directly by $f_n$ and varies with the token.

\paragraph{Associative scan.}
Despite its input-dependent retain and write terms, the recurrence preserves
the affine form required for an associative parallel scan. Define the
token-wise affine element
\begin{equation}
p_n
=
\left(
f_n,
w_n
\right),
\qquad
w_n
=
i_n
\odot
\left(
h_nB_n^{\top}
\right).
\end{equation}
Two consecutive recurrence elements compose according to
\begin{equation}
\left(
f_b,
w_b
\right)
\circ
\left(
f_a,
w_a
\right)
=
\left(
f_b\odot f_a,
\,
w_b
+
f_b\odot w_a
\right).
\label{eq:naju_scan_operator}
\end{equation}
This operator is associative because elementwise multiplication is associative
and distributive over addition. Consequently, the full state sequence can be
computed using a parallel prefix scan while producing the same result as the
sequential recurrence.

\subsection{Selective Parameterization}
\label{sec:selective_parameterization}

We now specify how the content drive, retain/write gates, and selective
write/read vectors are generated from the block input. Each Naju block first
projects the residual-stream input into two expanded branches:
\begin{equation}
\left(
\tilde{u}_n^{(l)},
z_n^{(l)}
\right)
=
\mathrm{split}
\left(
W_{\mathrm{in}}^{(l)}u_n^{(l)}
\right).
\label{eq:naju_input_projection}
\end{equation}
The first branch, $\tilde{u}_n^{(l)}$, generates the content drive used by the
recurrent update, whereas $z_n^{(l)}$ is reserved for output modulation in
Section~\ref{sec:block_output}.

The content branch is processed by a short depthwise causal convolution and a
SiLU nonlinearity:
\begin{equation}
h_n^{(l)}
=
\mathrm{SiLU}
\left(
\mathrm{DWConv}^{(l)}
\left(
\tilde{u}^{(l)}
\right)_n
\right),
\label{eq:naju_content_branch}
\end{equation}
where
\begin{equation}
\mathrm{SiLU}(x)
=
x\,\sigma(x).
\end{equation}
Because the convolution is depthwise and causal, each expanded channel
incorporates only its own recent history and no future positions. This gives
$h_n^{(l)}$ a short local context before it is used to generate the recurrent
write, gates, and selective vectors. Related local-mixing mechanisms have been
used in H3~\cite{fu2023h3} and Mamba~\cite{gu2023mamba} to expose nearby
token interactions before the recurrent update.

The retain and write decisions are generated from the shared content drive:
\begin{equation}
\begin{aligned}
a_n^{(l)}
&=
W_f^{(l)}h_n^{(l)}
+
\mathrm{DWConv}_f^{(l)}
\left(
h^{(l)}
\right)_n
+
f_{\mathrm{bias}}^{(l)}, \\
b_n^{(l)}
&=
W_i^{(l)}h_n^{(l)}
+
\mathrm{DWConv}_i^{(l)}
\left(
h^{(l)}
\right)_n
+
i_{\mathrm{bias}}^{(l)} .
\end{aligned}
\label{eq:naju_gate_logits}
\end{equation}
The corresponding gates are given in Eq.~(\ref{eq:naju_gates}).
The linear terms in Eq.~\eqref{eq:naju_gate_logits} transform the current
content representation, while the bias-free depthwise convolutions provide
each gate with an additional short history specific to that expanded channel.
The forget gate $f_n^{(l)}$ determines the token-dependent discrete pole, and
the input gate $i_n^{(l)}$ independently determines the write magnitude.

The selective write and read directions are generated from the same content
drive:
\begin{equation}
\left(
B_n^{(l)},
C_n^{(l)}
\right)
=
\mathrm{split}
\left(
W_{BC}^{(l)}h_n^{(l)}
\right).
\label{eq:naju_BC}
\end{equation}
Here, $B_n^{(l)}$ determines the state-space direction into which the current
content is written, whereas $C_n^{(l)}$ determines the direction along which
the recurrent state is read. Both vectors vary with the token and are shared
across expanded channels at that token.

Combining Eqs.~\eqref{eq:naju_content_branch}--\eqref{eq:naju_BC} with the
state update gives
\begin{equation}
x_n^{(l)}
=
f_n^{(l)}
\odot
x_{n-1}^{(l)}
+
i_n^{(l)}
\odot
\left(
h_n^{(l)}
\left(
B_n^{(l)}
\right)^{\top}
\right).
\end{equation}
Thus, Naju makes four token-dependent decisions from a shared locally informed
representation: how much state to retain, how strongly to write, where in the
state space to write, and how to read the resulting state. The recurrent
update nevertheless remains diagonal and affine, preserving compatibility with
an associative scan.

\subsection{Block Output}
\label{sec:block_output}

The recurrent output from Eq.~\eqref{eq:naju_readout} is combined with the
second branch of the input projection through multiplicative output
modulation:
\begin{equation}
u_n^{(l+1)}
=
W_o^{(l)}
\left(
y_n^{(l)}
\odot
\mathrm{SiLU}
\left(
z_n^{(l)}
\right)
\right)
+
u_n^{(l)} .
\label{eq:naju_block_output}
\end{equation}
The vector $z_n^{(l)}$ is generated together with the content branch in
Eq.~\eqref{eq:naju_input_projection}. After the SiLU nonlinearity, it
modulates the recurrent output elementwise before the output projection
$W_o^{(l)}$. This follows the GLU/SwiGLU-style multiplicative organization
used in modern sequence blocks
\cite{dauphin2017glu,shazeer2020glu}.

The term $\mathrm{SiLU}(z_n^{(l)})$ is more precisely an
\emph{output-modulation branch} than a probabilistic gate: unlike the sigmoid
gates $f_n^{(l)}$ and $i_n^{(l)}$, it is not restricted to the interval
$(0,1)$. Its role is to control which expanded-channel features of
$y_n^{(l)}$ are emitted through the output projection, without modifying the
recurrent state itself.

The block contains two nonrecurrent routes for current information. The first
is the feedthrough term inside the local readout,
\begin{equation}
D^{(l)}
\odot
h_n^{(l)},
\end{equation}
which is modulated and projected together with the recurrent-memory readout.
The second is the outer residual connection in
Eq.~\eqref{eq:naju_block_output}, which adds the original block input
$u_n^{(l)}$ directly to the projected sequence-mixer output. The feedthrough
therefore provides a learnable local route within the sequence mixer, whereas
the residual connection preserves a clean identity path across layers.

\paragraph{Complete block computation.}
Equations~\eqref{eq:naju_state_update}--\eqref{eq:naju_block_output} define
the complete Naju block used in the main experiments. Its computation can be
summarized as
\begin{equation}
u_n^{(l)}
\longrightarrow
\left(
h_n^{(l)},
z_n^{(l)}
\right)
\longrightarrow
\left(
f_n^{(l)},
i_n^{(l)},
B_n^{(l)},
C_n^{(l)}
\right)
\longrightarrow
x_n^{(l)}
\longrightarrow
y_n^{(l)}
\longrightarrow
u_n^{(l+1)} .
\end{equation}
The only temporally recurrent quantity is $x_n^{(l)}$. The projections,
depthwise convolutions, gates, selective vectors, readout, and output
modulation are token-wise or short-context feedforward computations around
that state.

Functionally, $f_n^{(l)}$ and $i_n^{(l)}$ provide retain and write control
analogous to the forget and input gates of an LSTM. The
$\mathrm{SiLU}(z_n^{(l)})$ branch provides gate-like output modulation, but
it is not an LSTM output gate and does not alter the recurrent state.
Consequently, Naju retains an affine diagonal state-space recurrence while
providing separate control over retention, writing, selective reading, and
block output.

\begin{figure*}[t]
\centering
\resizebox{\textwidth}{!}{%
\begin{tikzpicture}[
  font=\small,
  box/.style={draw,rounded corners=2pt,align=center,inner sep=2.5pt,minimum height=6.5mm},
  blk/.style={draw,rounded corners=2pt,fill=blue!6,align=center,inner sep=2.5pt,minimum height=6.5mm},
  op/.style={circle,draw,inner sep=0pt,minimum size=4.4mm},
  sig/.style={-{Latex[length=1.6mm]}},
]

\node (tok) {$w_{1:L}$};
\node[box,right=3.5mm of tok] (emb) {Embed};
\node[blk,right=3.5mm of emb] (b1) {block 1};
\node[blk,right=3.5mm of b1] (b2) {block 2};
\node[right=3.5mm of b2] (dots) {$\cdots$};
\node[blk,right=3.5mm of dots] (bL) {block $M$};
\node[box,right=3.5mm of bL] (norm) {RMSNorm};
\node[box,right=3.5mm of norm] (head) {head};
\node[right=3.5mm of head] (yh) {$\hat y$};

\foreach \a/\b in {tok/emb,emb/b1,b1/b2,b2/dots,dots/bL,bL/norm,norm/head,head/yh}
  \draw[sig] (\a) -- (\b);

\begin{scope}[yshift=-3.15cm]
  \node (uin)  at (0,0) {$u_n^{(l)}$};

  \node[box]  (inp)  at (1.55,0) {in-proj};

  \node[box]  (conv) at (3.3,-0.65) {Conv1d\\+ SiLU};

  \node[box]  (gates) at (5.5,-0.65)
    {$f_n=\sigma(a_n)$\\$i_n=\sigma(b_n)$\\$B_n,\,C_n$};

  \node[blk]  (scan) at (8.1,-0.65)
    {Naju junction\\parallel scan\\$\to x_n$};

  \node[box]  (read) at (11.2,-0.65)
 {readout\\$y_n=\alpha\,x_nC_n+D\odot h_n$};

  \node[box]  (zg) at (6.10,1.05)
    {SiLU$(z_n)$};

  \node[op]   (og)  at (13.35,-0.65) {$\otimes$};
  \node[box]  (wo)  at (14.45,-0.65) {$W_o$};
  \node[op]   (sum) at (15.55,-0.65) {$+$};
  \node (uout) at (16.70,-0.65) {$u_n^{(l+1)}$};

  \draw[sig] (uin) -- (inp);
  \draw[sig] (inp) |- (conv);
  \draw[sig] (conv) -- (gates);
  \draw[sig] (gates) -- (scan);
  \draw[sig] (scan) -- (read);
  \draw[sig] (read) -- (og);
  \draw[sig] (og) -- (wo);
  \draw[sig] (wo) -- (sum);
  \draw[sig] (sum) -- (uout);

  \draw[sig] (inp) |- (zg);
  \draw[sig] (zg.east) -| (og.north);
  \node[font=\scriptsize] at (11.8,0.80) {output gate};

  \draw[sig] (4.3,-0.65) -- ++(0,-0.90) -| (read.south);
  \node[font=\scriptsize] at (4.3,-0.44) {$h_n$};

  \draw[sig] (uin.south) -- ++(0,-1.95) -| (sum.south);
  \node[font=\scriptsize] at (6.,-2.0)
    {outer residual: carries $u_n^{(l)}$ forward};

  \coordinate (toppad) at (8.3,1.45);
  \coordinate (botpad) at (8.3,-2.35);
  \node[
    draw,dashed,rounded corners,
    fit=(uin)(zg)(scan)(read)(uout)(toppad)(botpad),
    inner sep=5pt
  ] (bb) {};
\end{scope}

\draw[dashed,gray] (b2.south west) -- (bb.north west);
\draw[dashed,gray] (b2.south east) -- (bb.north east);

\end{tikzpicture}}
\caption{
\textbf{Naju architecture.} The full model (top) stacks $M$ Naju blocks between a token embedding and a final prediction head; one block is expanded below. The block input $u_n^{(l)}$ is projected into a content branch $h_n^{(l)}$ and an output-modulation branch $z_n^{(l)}$. The content branch generates the token-dependent retain/write gates $f_n^{(l)}$ and $i_n^{(l)}$ and the selective write/read vectors $B_n^{(l)}$ and $C_n^{(l)}$, which define the native-discrete recurrence in Eq.~\eqref{eq:naju_state_update}. The resulting state is combined with the direct feedthrough path according to Eq.~\eqref{eq:naju_readout}. The $z_n^{(l)}$ branch modulates this local output before the output projection, and the outer residual connection preserves an identity path across blocks. Only $x_n^{(l)}$ is temporally recurrent; the remaining components are token-wise or short-context computations around the recurrent state.
}
\label{fig:block}
\end{figure*}
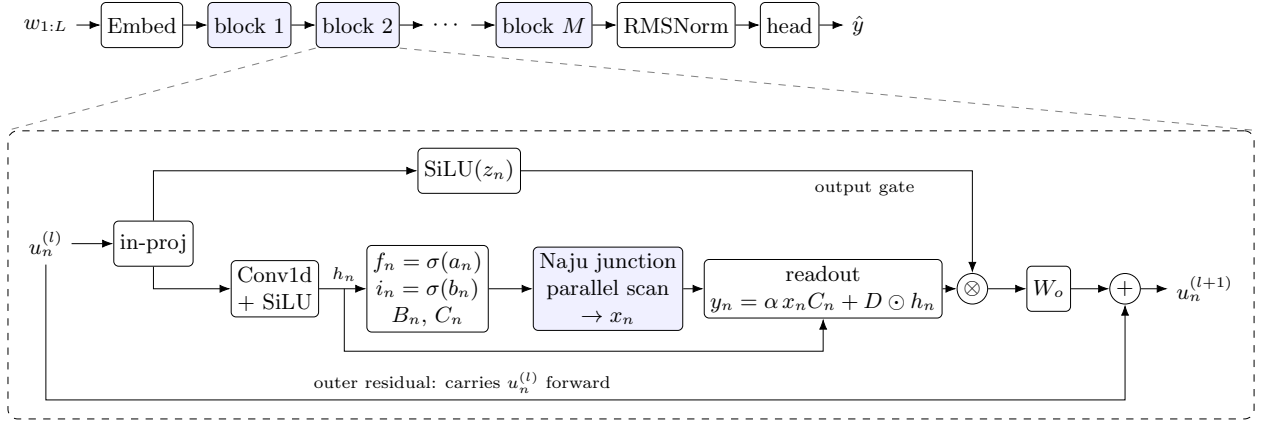

Figure~\ref{fig:block} shows the complete Naju block and the stacking of $M$
blocks into the full model. Each block consists of the native-discrete Naju
junction wrapped by input projection, short depthwise causal convolutions,
selective parameter generation, output modulation, output projection, and a
residual connection. The complete model otherwise follows a standard sequence
architecture with token embeddings, $M$ Naju blocks, RMSNorm, and a final
prediction head.

\section{Memory-Kernel and Pole-Gain Analysis}
\label{sec:kernel_pole_gain}

\subsection{Memory-Kernel View of Retain--Write Decoupling}
\label{sec:memory_kernel_decoupling}

For clarity, we omit the layer index and analyze one expanded channel. The
corresponding state update is
\begin{equation}
x_n
=
f_n x_{n-1}
+
i_n B_n h_n,
\label{eq:naju_time_varying_recurrence}
\end{equation}
where
$x_n,B_n,C_n
\in
\mathbb{R}^{d_{\text{state}}}$
and 
$f_n,i_n,h_n,D
\in
\mathbb{R}$.
The full recurrence applies the same analysis independently to all expanded
channels.

\paragraph{Unrolled memory dynamics.}
Unrolling Eq.~\eqref{eq:naju_time_varying_recurrence} gives
\begin{equation}
x_n
=
\left(
\prod_{j=1}^{n} f_j
\right)
x_0
+
\sum_{m=1}^{n}
\left[
\left(
\prod_{j=m+1}^{n} f_j
\right)
i_m B_m h_m
\right],
\label{eq:naju_unrolled_kernel}
\end{equation}
where an empty product is defined as one. Define the post-write retention
factor
\begin{equation}
R_{m\to n}
=
\prod_{j=m+1}^{n} f_j,
\qquad
R_{n\to n}=1.
\label{eq:naju_retention_product}
\end{equation}
The contribution written at position $m$ and remaining in the state at
position $n$ is therefore
\begin{equation}
x_{m\to n}
=
R_{m\to n}\,i_m B_m h_m.
\label{eq:naju_token_contribution}
\end{equation}
This expression separates the roles of the recurrent components. The drive
$h_m$ supplies the content amplitude, $B_m$ determines the state-space
direction of the write, $i_m$ controls how strongly the content is written,
and $R_{m\to n}$ determines how much of that write survives until position
$n$.

\paragraph{Output memory kernel.}
For the same channel, the readout is
\begin{equation}
y_n
=
\alpha\,C_n^{\top}x_n
+
Dh_n.
\label{eq:naju_readout_kernel}
\end{equation}
Substituting Eq.~\eqref{eq:naju_unrolled_kernel} yields
\begin{equation}
y_n
=
\alpha
\left(
\prod_{j=1}^{n} f_j
\right)
C_n^{\top}x_0
+
\sum_{m=1}^{n}
\alpha\,
R_{m\to n}\,
i_m\,
C_n^{\top}B_m\,
h_m
+
Dh_n.
\label{eq:naju_output_kernel}
\end{equation}
The causal memory kernel from position $m$ to position $n$ is thus
\begin{equation}
K_{n,m}
=
\begin{cases}
\alpha\,
R_{m\to n}\,
i_m\,
C_n^{\top}B_m,
&
m\le n, \\
0,
&
m>n.
\end{cases}
\label{eq:naju_memory_kernel}
\end{equation}
Accordingly,
\begin{equation}
y_n
=
\alpha
\left(
\prod_{j=1}^{n} f_j
\right)
C_n^{\top}x_0
+
\sum_{m=1}^{n}
K_{n,m}h_m
+
Dh_n.
\label{eq:naju_kernel_form}
\end{equation}
For a past token $m<n$, its contribution reaches the output only through the
recurrent-memory path:
\begin{equation}
y_{m\to n}
=
K_{n,m}h_m.
\label{eq:naju_past_output_contribution}
\end{equation}
For the current token, $R_{n\to n}=1$, so the output contains both the
state-mediated write and the direct feedthrough:
\begin{equation}
y_{n\to n}
=
\left(
\alpha\,i_n C_n^{\top}B_n
+
D
\right)
h_n.
\label{eq:naju_current_output_contribution}
\end{equation}

\paragraph{Retain--write decoupling.}
The kernel in Eq.~\eqref{eq:naju_memory_kernel} makes the retain--write
separation explicit. The token-dependent retention coefficient $f_j$
determines the survival product
\begin{equation}
R_{m\to n}
=
\prod_{j=m+1}^{n}f_j,
\end{equation}
whereas $i_m$ controls the write amplitude at the moment token $m$ is stored.
The vectors $B_m$ and $C_n$ determine the state-space write and read
directions, respectively, and $\alpha$ controls the overall scale of the
memory-mediated output.
This survival product also connects the kernel view to the local pole
interpretation. If the retention coefficients are frozen to a constant
$f_j=f$ after a write at position $m$, then
\begin{equation}
R_{m\to n}
=
f^{\,n-m}.
\end{equation}
Equivalently, the stored contribution from position $m$ evolves according to
\begin{equation}
x_{m\to n}
=
f\,x_{m\to n-1},
\qquad
n>m,
\end{equation}
whose transfer denominator is $1-fz^{-1}$ and whose discrete-time pole is
located at $z=f$. In the full time-varying recurrence, $f_n$ can therefore be
interpreted as a token-dependent local retention factor, while products of
these factors determine how long each stored contribution survives. The
direct feedthrough term $Dh_n$ affects only the instantaneous output and does
not alter either the recurrent retention dynamics or the retain--write
decomposition.

The retain--write decomposition acts directly on a selective SSM-style state:
$f_n$ controls temporal retention, $i_nB_nh_n$ defines the selective write,
and $C_n$ determines how the stored state is queried. Unlike
continuous-time-parameterized selective SSMs, Naju does not obtain its
retention factor through $\exp(\Delta_n A)$ or another discretization map.
Instead, the discrete retention coefficient is parameterized directly by
$f_n$, while the recurrence remains affine, diagonal, and compatible with an
associative scan.

\paragraph{Complementary single-gate coupling.}
The benefit of independent retain and write gates can be seen by considering
a complementary single-gate recurrence,
\begin{equation}
x_n
=
(1-g_n)\lambda_n x_{n-1}
+
g_n v_n,
\qquad
0<g_n<1,
\label{eq:coupled_gate}
\end{equation}
where $\lambda_n$ is an underlying transition coefficient and $v_n$ is the
candidate write. Its effective retention and write coefficients are
\begin{equation}
r_n
=
(1-g_n)\lambda_n,
\qquad
w_n
=
g_n.
\label{eq:coupled_rw}
\end{equation}
If the underlying transition is non-expansive,
$|\lambda_n|\le 1$, then
\begin{equation}
|r_n|+w_n
=
(1-g_n)|\lambda_n|+g_n
\le
(1-g_n)+g_n
=
1.
\label{eq:coupled_bound}
\end{equation}
Consequently,
\begin{equation}
|r_n|
\ge
1-\epsilon
\quad\Longrightarrow\quad
w_n
\le
\epsilon,
\label{eq:retain_implies_weak_write}
\end{equation}
and
\begin{equation}
w_n
\ge
1-\epsilon
\quad\Longrightarrow\quad
|r_n|
\le
\epsilon.
\label{eq:write_implies_weak_retain}
\end{equation}
Thus, a complementary gate can alternate between retaining and updating, but
it cannot choose the two gains independently at the same step.

Naju removes this structural constraint through
$
x_n
=
f_nx_{n-1}
+
i_nB_nh_n$, 
for which
$r_n=f_n$
and 
$w_n=i_n$.

Because $f_n$ and $i_n$ are generated independently, near-unit retention does
not impose an upper bound on the write gate.

\paragraph{Relation to the gated-RNN limit of selective SSMs.}
The connection to selective SSMs is formal rather than purely analogical.
Theorem~1 of Mamba~\cite{gu2023mamba} shows that, under the scalar
specialization
\begin{equation}
d_{\text{state}}=1,
\qquad
A=-1,
\qquad
B=1,
\end{equation}
the ZOH-discretized selective recurrence reduces to
\begin{equation}
s_n
=
(1-g_n)s_{n-1}
+
g_nu_n,
\qquad
g_n
=
\sigma\!\left(
\mathrm{Linear}(u_n)
\right).
\label{eq:mamba_gated_rnn_limit}
\end{equation}
In this specialization, the same gate controls the complementary pair
\begin{equation}
\text{retention}=1-g_n,
\qquad
\text{write}=g_n,
\end{equation}
which is precisely the coupling characterized in
Eqs.~\eqref{eq:coupled_bound}--\eqref{eq:write_implies_weak_retain}.

Naju starts from a directly parameterized discrete recurrence rather than
obtaining the gate through ZOH discretization, and replaces the complementary
pair $(1-g_n,g_n)$ with the independent pair $(f_n,i_n)$. The scalar
gated-RNN specialization is recovered by setting
\begin{equation}
f_n=1-g_n,
\qquad
i_n=g_n,
\qquad
B_n=1,
\qquad
h_n=u_n.
\end{equation}
Naju relaxes these constraints while retaining the input-dependent selective
write and read directions $B_n$ and $C_n$.

This comparison concerns the gated-RNN specialization of the selective SSM,
rather than the full expressive class of Mamba-style models. It identifies
the precise point of contact: Naju preserves the selective SSM state and
scan-compatible affine structure while replacing complementary retention and
writing with independently controlled native-discrete gains.

\subsection{Frozen-Gate Transfer Function}
\label{sec:frozen_transfer}

The native-discrete form of Naju admits a simple local transfer-function
interpretation. Although the complete layer is input dependent and therefore
time varying, its behavior can be analyzed locally by freezing the gates and
selective parameters over a short region.
For one expanded channel and one state coordinate, the locally frozen dynamics
reduce to the scalar recurrence
\begin{equation}
x[n]
=
f x[n-1]
+
iB h[n],
\qquad
y[n]
=
\alpha\,C x[n]
+
D h[n],
\label{eq:naju_scalar_recurrence}
\end{equation}
where $f$, $i$, $B$, $C$, and $D$ are treated as constants within the local
region. The scalar $h[n]$ denotes the corresponding component of the convolved
content drive $h_n^{(l)}$. The term $Dh[n]$ is a direct feedthrough term and
does not enter the recurrent state update.

\paragraph{Transfer function.}
Taking the $z$-transform of Eq.~\eqref{eq:naju_scalar_recurrence} under zero
initial state gives
\begin{equation}
X(z)
=
fz^{-1}X(z)
+
iB H_{\mathrm{in}}(z),
\end{equation}
where $H_{\mathrm{in}}(z)$ denotes the transform of $h[n]$. Solving for the
state gives
\begin{equation}
X(z)
=
\frac{iB}{1-fz^{-1}}
H_{\mathrm{in}}(z).
\end{equation}
Note that the transform variable $z$ used above is the standard $z$-domain variable and
is unrelated to the output-modulation branch $z_n^{(l)}$ in
Eq.~\eqref{eq:naju_input_projection}.
The local transfer function from the content drive to the readout is therefore
\begin{equation}
\mathcal{H}(z)
=
\frac{Y(z)}{H_{\mathrm{in}}(z)}
=
D
+
\frac{\alpha\,iCB}{1-fz^{-1}}
=
\frac{D(1-fz^{-1})+\alpha\,iCB}
     {1-fz^{-1}}.
\label{eq:naju_transfer_function}
\end{equation}
Equivalently, multiplying the numerator and denominator by $z$ gives
\begin{equation}
\mathcal{H}(z)
=
\frac{(D+\alpha\,iCB)z-Df}
     {z-f}.
\label{eq:naju_transfer_function_z}
\end{equation}
When the direct feedthrough path is removed, the transfer function reduces to
\begin{equation}
\mathcal{H}_{D=0}(z)
=
\frac{\alpha\,iCB}
     {1-fz^{-1}}
=
\frac{\alpha\,iCB\,z}
     {z-f}.
\label{eq:naju_transfer_function_no_D}
\end{equation}

\paragraph{Pole and gain interpretation.}
Equations~\eqref{eq:naju_transfer_function} and
\eqref{eq:naju_transfer_function_z} show that each locally frozen Naju
coordinate behaves as a first-order discrete IIR filter with a direct
feedthrough path. Its recurrent pole is located at
\begin{equation}
z=f.
\end{equation}
Thus, the forget gate determines the local memory dynamics. The input gate
$i$ controls the write gain, while $B$ and $C$ determine the local write and
read gains. The scalar $\alpha$ controls the overall scale of the
state-mediated readout.

The direct term $D$ modifies the numerator and therefore the zero structure of
the transfer function. It does not modify the recurrent pole because it does
not enter the state recurrence. Except in degenerate cases where the recurrent
term vanishes or an exact pole--zero cancellation occurs, the pole remains at
$z=f$.

\paragraph{Memory time scale.}
Because
\begin{equation}
f
=
\sigma(a)
\in
(0,1)
\end{equation}
for finite logits, the locally frozen pole lies inside the unit circle. A
stored contribution is multiplied by $f^k$ after $k$ recurrent steps. When
$f$ approaches one, the pole approaches $z=1$, producing a near-integrator
regime in which stored information decays slowly over finite horizons. When
$f$ is small, the pole approaches the origin and the recurrent state rapidly
forgets past content.

A natural summary of this decay rate is the e-folding memory time constant
\begin{equation}
\tau_e(f)
=
\frac{1}{-\log f}
\approx
\frac{1}{1-f},
\qquad
f\to 1.
\label{eq:tau_e}
\end{equation}
After approximately $\tau_e(f)$ steps, the magnitude of a stored contribution
has decayed to $
e^{-1}
\approx
0.37
$ of its original value.
For example, if $f=0.999$, then $\tau_e\approx 999.5$ steps. A stored
contribution retains approximately $36.8\%$ of its original magnitude after
$1{,}000$ steps and $13.5\%$ after $2{,}000$ steps. 

Thus, a pole close to one supports long but gradually decaying memory; the
time constant is not a finite horizon at which the stored contribution
abruptly vanishes. This example describes the locally frozen dynamics, while
the full Naju model uses token-dependent values of $f_n$ whose products
determine the actual retention over a sequence.

\subsection{Stability as a Consequence}
\label{sec:stability_consequence}

The pole and memory-kernel views make stability a direct consequence of the
Naju parameterization.
Under the locally frozen analysis of Section~\ref{sec:frozen_transfer}, the
only recurrent pole of each scalar coordinate is $z=f$.
Because the forget gate is parameterized as
\begin{equation}
f=\sigma(a)\in(0,1)
\end{equation}
for every finite logit $a$, each locally frozen coordinate is Schur stable:
its recurrent pole lies strictly inside the unit circle.
The direct feedthrough term $Dh[n]$ affects only the instantaneous readout
and the numerator of the transfer function. Since it does not enter the state
recurrence, it cannot move the recurrent pole or alter recurrent stability.

For the full time-varying recurrence, consider one scalar state coordinate,
\begin{equation}
x_n
=
f_nx_{n-1}
+
v_n,
\qquad
v_n
=
i_nB_nh_n.
\label{eq:stability_scalar_recurrence}
\end{equation}
Pointwise finiteness of the forget logits is not sufficient to guarantee a
uniform stability bound, because the logits could remain finite at each step
while approaching infinity along the sequence. We therefore assume that the
forget logits are uniformly upper bounded:
\begin{equation}
a_n
\le
A_{\max}
<
\infty
\qquad
\text{for all } n.
\end{equation}
This gives
$
0
<
f_n
=
\sigma(a_n)
\le
\rho
:=
\sigma(A_{\max})
<
1$.
The retention product in Eq.~\eqref{eq:naju_retention_product} then satisfies
\begin{equation}
R_{m\to n}
=
\prod_{j=m+1}^{n}f_j
\le
\rho^{\,n-m}.
\end{equation}
Thus, the influence of a write at position $m$ decays geometrically with its
distance from position $n$. If the write drive and readout quantities are also
uniformly bounded, the recurrent state and output remain bounded.

\begin{theorem}[Fading-memory/BIBO bound under uniformly bounded gates]
\label{thm:bibo}
Consider one scalar coordinate of the Naju recurrence $x_n=f_nx_{n-1}+v_n$ with
$f_n=\sigma(a_n)$ and $v_n=i_n(B_nh_n)$. Suppose there exist finite constants
$A_{\max}$, $M_v$, $M_C$, $M_D$, and $M_h$ such that $a_n\le A_{\max}$,
$|v_n|\le M_v$, $|C_n|\le M_C$, $|D|\le M_D$, and $|h_n|\le M_h$ for all $n$.
Then, with $\rho=\sigma(A_{\max})<1$ and the constant readout gain $\alpha>0$,
\[
|x_n|\;\le\;\rho^n |x_0|+\frac{M_v}{1-\rho},
\qquad
|y_n|\;\le\;\alpha M_C\Big(\rho^n |x_0|+\frac{M_v}{1-\rho}\Big)+M_DM_h ,
\]
so the coordinate satisfies a fading-memory/BIBO bound with no stability
regularizer; the diagonal vector case follows coordinatewise.
\end{theorem}

This property does not require an additional stability regularizer. It follows
from using the sigmoid forget gate directly as the discrete retention
coefficient. The pole can approach the unit circle arbitrarily closely,
supporting near-lossless retention over a finite horizon, but it cannot equal
or exceed one for finite logits. Naju is therefore stable without being
restricted to an exact lossless integrator. A proof of
Theorem~\ref{thm:bibo} is provided in
Appendix~\ref{app:stability_bound}.

\paragraph{Readout gain and dimension correction.}
Theorem~\ref{thm:bibo} also clarifies the role of the readout gain $\alpha$.
Because $\alpha$ appears only in the output equation, it does not affect the
recurrent pole, the state trajectory, or recurrent stability. Instead, it
controls the scale and optimization sensitivity of the state-readout pathway.

For one expanded channel, the recurrent contribution is
\begin{equation}
r_n
=
C_n^{\top}x_n
=
\sum_{k=1}^{d_{\text{state}}}
C_{n,k}x_{n,k}.
\end{equation}
Under the standard initialization assumption that the coordinate-wise terms
have comparable scale and are approximately uncorrelated, the root-mean-square
magnitude of this sum grows as
\begin{equation}
\operatorname{RMS}
\left(
C_n^{\top}x_n
\right)
=
\mathcal{O}
\left(
\sqrt{d_{\text{state}}}
\right).
\end{equation}
Without correction, increasing $d_{\text{state}}$ can therefore enlarge the
state-mediated output and its corresponding gradients even though the
recurrent dynamics themselves remain unchanged.

We compensate for this width-dependent readout scale by using
\begin{equation}
\alpha
=
\frac{1}{\sqrt{d_{\text{state}}}}.
\label{eq:readout_norm}
\end{equation}
This parameter-free normalization cancels the leading
$\sqrt{d_{\text{state}}}$ dependence and keeps the state-readout pathway on a
comparable scale as the state dimension changes. Unlike a global learning-rate
adjustment, $\alpha$ specifically rescales the recurrent readout relative to
the direct and residual pathways.

Unless otherwise stated, Eq.~\eqref{eq:readout_norm} is used as the
dimension-corrected readout gain in the main experiments. This choice is an
optimization and scale-normalization convention rather than a stability
requirement: Theorem~\ref{thm:bibo} holds for any fixed $\alpha>0$.

\section{Experiments}
\label{sec:experiments}

\subsection{Diagnostic Suite for Retention--Overwrite Trade-offs}
\label{sec:suite}

We evaluate Naju with a controlled diagnostic suite designed to expose a
central tension in long-sequence state models: the same recurrent state must
sometimes preserve old information over long distances, but at other times must
overwrite stale information with newer content. The core suite consists of four
synthetic long-sequence tasks, labeled \textbf{T1}--\textbf{T4}. These tasks
cover associative recall, length extrapolation, and state tracking: T1 and T2
are key-value recall probes in the spirit of associative recall / multi-query
recall benchmarks for efficient sequence models
\cite{arora2023zoology,arora2024based}; and
T3 and T4 evaluate current-state tracking, following the motivation that sequence
models should maintain and update latent states rather than only retrieve fixed
facts \cite{liu2023transformers}.

The four tasks are deliberately organized into two groups. \textbf{T1} and
\textbf{T2} measure \emph{retention}: facts are written once and never updated, so the
model is rewarded for preserving bindings over increasingly long distances.
\textbf{T3} and \textbf{T4} measure \emph{overwriting}: an entity receives repeated
updates, and only the latest value is correct, so the model must suppress stale
answers and replace old content with new content. Thus, the task suite probes
whether a model can balance long-range preservation with selective replacement,
rather than optimizing only one side of the trade-off.

\begin{itemize}
    \item[\textbf{T1}] \textbf{Key-value retrieval.}
    A sequence contains several facts of the form
    \texttt{(KEY entity IS value SEP)}, followed by a query
    \texttt{(QUES KEY entity EOS)}. The label is the queried entity's value,
    and the queried fact is placed in an early, middle, or late region of the
sequence for distance-resolved evaluation for
    distance-resolved accuracy. This task probes basic associative recall.
    \emph{Example:}
    \[
    \texttt{KEY a IS 3 SEP KEY b IS 7 SEP QUES KEY a EOS}
    \rightarrow
    \texttt{3}.
    \]

    \item[\textbf{T2}] \textbf{Scattered key-value retrieval.}
    This task has the same semantic structure as T1, but facts are scattered
    across the sequence behind padding gaps. As the sequence length increases,
    the query-to-fact distance also increases. The model is trained on short
    sequences and evaluated on longer sequences, making T2 the primary
    length-extrapolation probe for retention.
    \emph{Example:}
    \[
    \texttt{KEY a IS 3 SEP}
    \ \langle\text{pad}\times k\rangle\
    \texttt{KEY b IS 7 SEP}
    \ \langle\text{pad}\times k\rangle\
    \texttt{QUES KEY a EOS}
    \rightarrow
    \texttt{3},
    \]
    where the gap $k$, and therefore the distance from the queried fact to the
    query, grows at test time.

    \item[\textbf{T3}] \textbf{Current-state tracking.}
    An entity receives several updates among distractors, and the model must
    report the latest valid value. Each update has the form
    \texttt{(CONN entity LIKES value SEP)}, where the connective \texttt{CONN}
    is \texttt{INIT}, \texttt{LATER}, or \texttt{FINAL} for the target entity's
    first, middle, or last update. The query is
    \texttt{(QUES entity NOW EOS)}. This task measures stale-answer suppression
    and basic overwriting.
    \emph{Example:}
    \[
    \texttt{INIT x LIKES 3 SEP}\ \ldots\
    \texttt{LATER x LIKES 5 SEP}\ \ldots\
    \texttt{FINAL x LIKES 8 SEP}\ \ldots\
    \texttt{QUES x NOW EOS}
    \rightarrow
    \texttt{8}.
    \]
    Here \texttt{3} and \texttt{5} are stale values, and the model must return
the value associated with the latest valid update.

    \item[\textbf{T4}] \textbf{Recency-only current-state tracking.}
    This task hardens T3 along several axes. First, the final-update marker is
    removed: every target update uses the same \texttt{LATER} connective, so
    the latest value must be identified by recency rather than by
    pattern-matching a \texttt{FINAL} token. Second, the target's updates are
    scattered across the sequence, so the distance from the latest update to
    the query grows. Third, the value pool is enlarged, increasing the output
    space. Finally, the model is trained on short sequences and evaluated on
    longer sequences. T4 is therefore the strongest stress test for overwriting
    under length extrapolation.
    \emph{Example:}
    \[
    \texttt{LATER x LIKES 3 SEP}\ \ldots\
    \texttt{LATER x LIKES 5 SEP}\ \ldots\
    \texttt{LATER x LIKES 8 SEP}\ \ldots\
    \texttt{QUES x NOW EOS}
    \rightarrow
    \texttt{8},
    \]
    where \texttt{8} must be identified as the most recent write by recency
    alone, at test-time distances beyond those seen during training.
\end{itemize}

All four tasks are evaluated as single-answer classification rather than
autoregressive generation. The model reads the full sequence, and a linear head
applied to the final \texttt{EOS} position predicts one value from the value
vocabulary. A prediction is counted as correct when it matches the sampled label.
Accuracy is measured over freshly sampled held-out sequences with unseen
entity--value bindings and is also reported by distance bucket when applicable.
For length-extrapolation results, the model is trained at length $512$ and
tested at longer lengths such as $1024$ and $2048$, so the relevant fact or
latest update lies farther from the query than any dependency observed during
training.

Together, the four tasks make the retention--overwrite trade-off explicit:
a strong preservation bias is expected to favor T1/T2 but may retain stale
values in T3/T4, whereas an aggressive update bias may produce the opposite
trade-off---the tension that motivates
the decoupled Naju junction, where $f_n$ controls retention through the
memory kernel and $i_n$ the strength of new writes. The following
experiments therefore evaluate not only whether a model can remember, but
whether it can balance remembering with selective overwriting.

\subsection{Compared Models and Training Setup}
\label{sec:setup}

We compare Naju against a Transformer encoder, standard Mamba with the
official selective-scan implementation, Mamba-2, GLA, RetNet, xLSTM, RWKV, and
HGRN. Unless otherwise noted, models use $d_{\text{model}}=128$, $4$ layers,
and expansion factor $2$. The standard Mamba baseline uses $d_{\text{model}}=256$ and
$d_{\text{state}}=16$, following the configuration used in the reference
implementation for this benchmark. Because this gives Mamba a different model
width from the other baselines, we report parameter counts alongside all main
results. For SSM-style models with selective state maps, the write and read parameters are
input-dependent where applicable.

The \textbf{Naju configuration} uses $d_{\text{model}}=128$, $4$ layers,
expansion factor $2$, and $d_{\text{state}}=64$, with the readout exactly as
in Section~\ref{sec:readout}: the feedthrough $D$ is a learned
per-channel constant, initialized to the small value
$D_{\text{init}}=0.01$ so that the direct route starts nearly silent and the
recurrent memory path dominates early training, and the readout gain is
fixed at the parameter-free $\alpha=1/\sqrt{d_{\text{state}}}$
(Eq.~\eqref{eq:readout_norm}), which compensates for the leading
$\sqrt{d_{\text{state}}}$ growth of the state-readout scale under the
initialization assumptions of Section~\ref{sec:stability_consequence}.

For the Naju gates, we use a preserve-first initialization. Specifically, the
per-channel gate biases $f_{\mathrm{bias}}^{(l)}$ and $i_{\mathrm{bias}}^{(l)}$ of
Eq.~\eqref{eq:naju_gate_logits} are initialized to the constants $b_f$ and $b_i$
(the two knobs swept in Section~\ref{sec:ablations}),
\begin{equation}
b_f=+5,
\qquad
b_i=-2, \notag
\end{equation}
so that initially $\sigma(b_f)\approx 0.993$ and $\sigma(b_i)\approx 0.12$.
Thus, before learning input-specific deviations, Naju is strongly biased
toward retaining existing memory and writing conservatively. This initialization encourages long-range retention at the start of training
while leaving both gates free to learn token-dependent retain and write
behavior. The same initialization is used across all tasks.

\paragraph{Training protocol.}
All compared models use the shared optimization protocol summarized in
Table~\ref{tab:protocol}; architecture-specific settings are stated explicitly. No stability regularizer, auxiliary loss, or
task-specific regularizer is added. Each model is trained with the task loss
only. Headline results are reported as mean$\pm$std over $5$ seeds at a matched
training budget; remaining numbers are single-seed results where noted.

\begin{table}[t]
\centering\small
\begin{tabular}{ll}
\toprule
\textbf{Setting} & \textbf{Value} \\
\midrule
Optimizer & AdamW \\
Learning rate & $3\times10^{-4}$ \\
Weight decay & $10^{-4}$ \\
Batch size & $32$ \\
LR schedule & cosine with linear warmup \\
Warmup ratio & $0.05$ of total steps \\
Epochs & $50$ \\
Gradient clipping & $1.0$ \\
Train sequence length & $512$ \\
\bottomrule
\end{tabular}
\caption{
Shared training protocol. Every compared model is trained with these shared
optimization settings; only the architecture and explicitly stated per-model
hyperparameters differ. The standard Mamba baseline uses
$d_{\text{model}}=256$ and $d_{\text{state}}=16$ as stated in the text.
}
\label{tab:protocol}
\end{table}

\paragraph{Compute environment.}
Experiments are run on a single NVIDIA RTX~5090 (32\,GB) using PyTorch~2.7.1
with bfloat16 autocast. The PyTorch build uses CUDA~12.6, while the installed
driver supports CUDA~12.8. Unless otherwise stated, runtime and memory
measurements are collected on an otherwise idle GPU after warm-up and are
reported over repeated runs using the same model configurations as the
corresponding result tables. All completed runs use a single GPU; configurations that exceed the available memory are reported explicitly as OOM.

\subsection{Retention versus Overwriting}
\label{sec:diag_summary}
The four diagnostic tasks separate into two \emph{saturated} axes and two
\emph{contested} ones. On associative recall (T1) and current-state tracking (T3)
\emph{at the training length}, every capable model reaches ${\approx}100\%$, so these
tasks do not discriminate among the strong models. The discrimination comes from
\emph{length extrapolation} on the two opposing demands a memory places on the
state: T2 (scattered key-value) isolates near-lossless \emph{retention}---bindings never
change, so holding everything is optimal---while T4 (recency-only current-state tracking) isolates
active \emph{overwriting}---an entity is rewritten repeatedly and only the latest value
counts, so stale state must be forgotten. Table~\ref{tab:summary} reports all four tasks
at the longest evaluated length ($2048$, $4\times$ the training distance); the per-length
breakdowns and per-model analysis are in Appendix~\ref{app:pertask}.

\begin{table}[h]
\centering
\small
\begin{tabular}{lllcccccc}
\toprule
\textbf{Model} & \textbf{Config.} & \textbf{Memory} & \textbf{Params} & \textbf{State}$^{\S}$ & \textbf{T1} & \textbf{T2}$^{\ddagger}$ & \textbf{T3} & \textbf{T4}$^{\ddagger}$ \\
\midrule
Transformer  & $d\,256$, $h\,4$   & KV cache & 4.29M & $\mathcal{O}(L)$ & $0.11$ & $0.12$ & $0.06$ & $0.07$ \\
\;{\scriptsize$\hookrightarrow$ train\,/\,val @512} & & & & & {\scriptsize$0.72/0.13$} & {\scriptsize$0.99/0.13$} & {\scriptsize$0.73/0.72$} & {\scriptsize$0.71/0.12$} \\
Transformer ($150$ ep.) & $d\,256$, $h\,4$ & KV cache & 4.29M & $\mathcal{O}(L)$ & $0.11$ & $0.11$ & $0.06$ & $0.06$ \\
\;{\scriptsize$\hookrightarrow$ train\,/\,val @512} & & & & & {\scriptsize$0.98/0.14$} & {\scriptsize$1.00/0.13$} & {\scriptsize$0.93/0.92$} & {\scriptsize$1.00/0.13$} \\
HGRN         & $d\,128$           & scalar   & 0.29M & $128$      & $0.11$ & $0.06$ & $1.00$ & $0.08$ \\
GLA          & $d\,128$, $h\,4$   & matrix   & 0.31M & $2{,}048$  & $1.00$ & $0.07$ & $1.00$ & $\mathbf{0.88}$ \\
RWKV         & $d\,128$, $h\,4$   & matrix   & 0.51M & $2{,}048$  & $0.10$ & $0.06$ & $0.95$ & $0.08$ \\
RetNet       & $d\,128$, $h\,4$   & matrix   & 0.56M & $8{,}192$  & $1.00$ & $0.30$ & $0.74$ & $0.50$ \\
xLSTM        & $d\,128$, $h\,4$   & matrix   & 0.46M & $16{,}384$ & $1.00$ & $\mathbf{1.00}$ & $0.95$ & $0.73$ \\
Mamba        & $d\,256$, $N\,16$  & vector   & 1.84M & $8{,}192$  & $1.00$ & $0.62$ & $1.00$ & $0.77$ \\
Mamba-2      & $d\,128$, $N\,64$  & vector   & 0.50M & $16{,}384$ & $1.00$ & $0.33$ & $0.99$ & $0.68$ \\
Naju (ours)  & $d\,128$, $N\,64$  & vector   & 1.09M & $16{,}384$ & $1.00$ & $\mathbf{0.99}$ & $0.94$ & $\mathbf{0.89}$ \\
\bottomrule
\end{tabular}
\caption{\textbf{Summary of the diagnostic suite.}
Main rows report length-$2048$ accuracy after training at length $512$
(mean over $5$ seeds). The indented Transformer rows report
final-epoch train / best-validation accuracy at length $512$; the
$150$-epoch run uses three times the standard training budget.
The $\ddagger$ columns denote the contested retention and overwriting axes.
Bold marks the two highest mean accuracies on each contested axis, T2 and T4. Model configuration abbreviations and
per-layer state sizes are defined in the text; full per-length results and
standard deviations are reported in Appendix~\ref{app:pertask} (Tables~\ref{tab:main}--\ref{tab:sthard}).}
\label{tab:summary}
\end{table}

\paragraph{The two axes split the baselines.}
On retention-oriented T2, Naju ($0.99$) and xLSTM ($1.00$) remain near
perfect at length $2048$. In contrast, the Mamba baselines degrade as the
padded retrieval distance increases (Mamba $0.62$; Mamba-2 $0.33$), while
GLA, RWKV, and HGRN fall near chance despite stronger performance on the
contiguous T1 task. On overwrite-oriented T4, the ranking changes
substantially: Naju ($0.89$), GLA ($0.88$), Mamba ($0.77$), and xLSTM
($0.73$) perform best.

The contrast across the two axes is informative. xLSTM provides nearly
perfect retention but weaker overwriting ($1.00/0.73$), whereas GLA performs
strongly on overwriting but fails to retain information across long padded
distances ($0.07/0.88$). Naju remains strong on both axes ($0.99/0.89$) and
achieves the highest worst-axis accuracy,
$\min(\mathrm{T2},\mathrm{T4})=0.89$, among the compared models. This result
is consistent with the intended role of Naju's decoupled forget and input
gates: retention and writing can be adjusted independently rather than being
forced into a complementary trade-off.

The ranking is not explained by recurrent-state size alone. Mamba-2 matches
Naju's per-layer state size ($16{,}384$) but obtains $0.33$ on T2, and
increasing Mamba's state size does not improve performance in the matched
factorial experiment under the evaluated settings
(Table~\ref{tab:mamba_factorial}). Naju also uses approximately $1.09$M
parameters, compared with $1.84$M for Mamba.

\paragraph{Diagnosing the Transformer results.} The unusually low Transformer scores at length $2048$ are not explained by a single optimization failure. The training-length diagnostics reveal two distinct failure modes. On T1, T2, and T4, extending training from $50$ to $150$ epochs raises training accuracy to $0.98$--$1.00$, while validation accuracy at length $512$ remains near chance ($0.13$--$0.14$). Thus, the model can memorize the finite training set but does not learn the underlying entity--value rule, and tripling the training budget does not close this generalization gap. On T3, by contrast, training and validation accuracy rise together to $0.93$ and $0.92$, respectively, showing that the Transformer can fit and generalize within the training-length distribution. Its accuracy nevertheless falls to $0.06$ at length $2048$, indicating a separate failure of length extrapolation. These diagnostics support interpreting the low Transformer results as task-dependent generalization and extrapolation failures rather than insufficient convergence alone.

\subsection{Robustness of the Comparison: Mamba Width and State}
\label{app:mamba_factorial}

The main comparison evaluates Mamba using its reference configuration, $d_{\text{model}}=256$ and $d_{\text{state}}=16$, which has roughly twice as many parameters as the default Naju model. A natural concern is that Mamba's weaker performance on retention-oriented T2 and overwrite-oriented T4 may be specific to this width and state-size choice. To test this possibility, we evaluate the full factorial combination \begin{equation} d_{\text{model}} \in \{128,256\}, \qquad d_{\text{state}} \in \{16,32,64\}, \notag \end{equation} on the two contested tasks. Table~\ref{tab:mamba_factorial} reports length-$2048$ extrapolation accuracy over five seeds for T2 scattered key--value retrieval and T4 recency-only current-state tracking. We additionally include Naju configurations at the corresponding model widths and state sizes. T1 and T3 are omitted because the Mamba configurations evaluated in this sweep saturate those tasks at $1.00$ and therefore provide little discrimination.

\begin{table}[t]
\centering
\small
\begin{tabular}{llcc}
\toprule
\textbf{Config} & \textbf{Params} & \textbf{T2 @2048} & \textbf{T4 @2048} \\
\midrule
\multicolumn{4}{l}{\emph{Mamba} ($\exp(\Delta_n A)$ transition)} \\
$d\,128/N\,16$ & $0.55$M & $0.47${\scriptsize$\,\pm0.16$}\,{\scriptsize(min$\,0.31$)} & $0.90${\scriptsize$\,\pm0.11$}\,{\scriptsize(min$\,0.69$)} \\
$d\,128/N\,32$ & $0.60$M & $0.55${\scriptsize$\,\pm0.09$}\,{\scriptsize(min$\,0.48$)} & $0.68${\scriptsize$\,\pm0.20$}\,{\scriptsize(min$\,0.42$)} \\
$d\,128/N\,64$ & $0.70$M & $0.68${\scriptsize$\,\pm0.21$}\,{\scriptsize(min$\,0.40$)} & $0.76${\scriptsize$\,\pm0.34$}\,{\scriptsize(min$\,0.10$)} \\
$d\,256/N\,16$\,$^{\star}$ & $1.96$M & $0.62${\scriptsize$\,\pm0.30$}\,{\scriptsize(min$\,0.07$)} & $0.77${\scriptsize$\,\pm0.27$}\,{\scriptsize(min$\,0.29$)} \\
$d\,256/N\,32$ & $2.06$M & $0.67${\scriptsize$\,\pm0.24$}\,{\scriptsize(min$\,0.33$)} & $0.84${\scriptsize$\,\pm0.27$}\,{\scriptsize(min$\,0.31$)} \\
$d\,256/N\,64$ & $2.26$M & $0.73${\scriptsize$\,\pm0.33$}\,{\scriptsize(min$\,0.07$)} & $0.68${\scriptsize$\,\pm0.38$}\,{\scriptsize(min$\,0.10$)} \\
\midrule
\multicolumn{4}{l}{\emph{Naju} (ours)} \\
$d\,128/N\,32$ & $1.09$M & $0.99${\scriptsize$\,\pm0.00$}\,{\scriptsize(min$\,0.98$)} & $0.83${\scriptsize$\,\pm0.13$}\,{\scriptsize(min$\,0.70$)} \\
$d\,128/N\,64$ (default) & $1.15$M & $\mathbf{0.99}${\scriptsize$\,\pm0.00$}\,{\scriptsize(min$\,0.99$)} & $\mathbf{0.89}${\scriptsize$\,\pm0.06$}\,{\scriptsize(min$\,0.80$)} \\
$d\,256/N\,32$ & $4.04$M & $1.00${\scriptsize$\,\pm0.00$}\,{\scriptsize(min$\,1.00$)} & $0.78${\scriptsize$\,\pm0.23$}\,{\scriptsize(min$\,0.52$)} \\
$d\,256/N\,64$ & $4.18$M & $1.00${\scriptsize$\,\pm0.00$}\,{\scriptsize(min$\,0.99$)} & $0.80${\scriptsize$\,\pm0.24$}\,{\scriptsize(min$\,0.52$)} \\
\bottomrule
\end{tabular}
\caption{\textbf{Mamba width--state factorial on the two contested tasks.} Length-$2048$ extrapolation accuracy is reported as mean$\pm$std over seeds $1$--$5$, with the worst seed in parentheses. Row labels use the abbreviations of Table~\ref{tab:summary}. The $^{\star}$ row is the Mamba configuration used in the main comparison. Parameter counts differ slightly from Table~\ref{tab:summary} because T2 and T4 use larger task vocabularies. All Naju rows use the same final readout recipe at each width and state size.}
\label{tab:mamba_factorial}
\end{table}

\paragraph{Width--state robustness.} Three observations emerge from the factorial comparison.
First, Mamba's T2 performance varies only moderately across the evaluated
width and state configurations, ranging from $0.47$ to $0.73$, whereas all
Naju configurations remain near perfect at $0.99$--$1.00$. Thus, the
retention gap in the main comparison is not attributable to a single
unfavorable Mamba configuration.
Second, the configuration that performs best on one contested axis does not
necessarily perform best on the other. For example, Mamba
$d\,128/N\,16$ achieves strong mean T4 accuracy ($0.90$) but lower T2
accuracy ($0.47$), while larger-state configurations improve T2 only
partially and show greater seed variability on T4. None of the evaluated
Mamba configurations simultaneously matches default Naju on both axes.
Third, Naju remains strong across the tested state sizes. At
$d_{\text{model}}=128$, increasing $d_{\text{state}}$ from $32$ to $64$
preserves near-perfect T2 accuracy and raises mean T4 accuracy from $0.83$ to
$0.89$. The default configuration therefore provides the best observed
balance between retention and overwriting, achieving $0.99$ on T2 and $0.89$
on T4.

Overall, the factorial sweep confirms that Naju's joint retention--overwrite
performance is robust to the width and state-size choices used in the
comparison. The result supports the intended role of independently
parameterized retain and write gates without requiring a detailed
architecture-specific explanation of Mamba's behavior.

\subsection{Efficiency: Compute and Memory}
\label{sec:dataeff_sec}

\paragraph{Throughput and memory.}\label{sec:throughput}
We benchmark forward latency and peak memory at a matched width
($d_{\text{model}}=256$, $4$ layers, batch size $8$) over sequence lengths
$512$--$32768$ (Table~\ref{tab:throughput}). The Transformer is evaluated
using both standard math attention and flash/SDPA attention. The evaluated
recurrent and SSM implementations, excluding the parallel xLSTM implementation described below, exhibit approximately linear latency growth
with sequence length, whereas attention retains quadratic compute complexity.
Flash attention avoids materializing the full $L\times L$ attention matrix,
but its latency becomes comparable to matched-width Naju at length $4096$
($22.0$ vs.\ $21.5$ ms) and reaches $1{,}038$ ms at length $32768$, compared
with $173.3$ ms for Naju. Standard math attention and the evaluated xLSTM
implementation run out of memory at length $8192$. At length $4096$, xLSTM
uses $20{,}384$ MB, compared with $806$ MB for matched-width Naju.

Within the linear-scaling group, HGRN and GLA achieve the lowest measured
forward latency, while Mamba and Mamba-2 remain faster than the current
matched-width Naju implementation. Naju uses the native CUDA scan described
in Appendix~\ref{app:cuda_scan} and maintains linear latency scaling,
requiring $173.3$ ms at length $32768$, compared with $78.9$ ms for Mamba and
$52.9$ ms for Mamba-2. Thus, matched-width Naju is approximately $2.2\times$
slower than Mamba and $3.3\times$ slower than Mamba-2 at the longest measured
length. Its peak memory at the same length is $6{,}092$ MB, compared with
$4{,}968$ MB for Mamba and $3{,}988$ MB for Mamba-2. The additional memory is
consistent with Naju's independent gate activations and larger recurrent
state, although the benchmark does not isolate the contribution of each
component.

The larger parameter count of matched-width Naju is primarily attributable to
its independent retain and write parameterizations. In the current
implementation, the $f$- and $i$-gate branches each use a full
$d_{\text{inner}}\times d_{\text{inner}}$ projection, whereas Mamba uses a
lower-rank parameterization for its step-size branch. The rank-$32$ gate
ablation reduces Naju latency only modestly
(Section~\ref{sec:wt103}), suggesting that the scan execution path, rather
than the full-rank gate projections alone, accounts for most of the remaining
latency gap. This is an implementation-level difference rather than an
asymptotic one.

\begin{table}[t]
\centering\small
{\bfseries (a) Forward latency (ms\,/\,batch)}\\[2pt]
\begin{tabular}{lrc ccccccc}
\toprule
\textbf{Model} & \textbf{Params} & \textbf{Scaling} & \textbf{512} & \textbf{1024} & \textbf{2048} & \textbf{4096} & \textbf{8192} & \textbf{16k} & \textbf{32k} \\
\midrule
Transformer (math attn.)  & 11.81M & $\mathcal{O}(L^2)$ & 1.3 & 3.3 & 14.9 & 54.1 & --- & --- & --- \\
Transformer (flash attn.) & 11.81M & $\mathcal{O}(L^2)$ & 1.2 & 2.7 & 7.1 & 22.0 & 75.7 & 274.8 & 1{,}038 \\
HGRN & 1.32M & $\mathcal{O}(L)$ & 1.7 & 1.7 & 1.7 & 3.1 & 7.6 & 17.2 & 34.0 \\
GLA & 1.34M & $\mathcal{O}(L)$ & 2.7 & 2.6 & 2.6 & 3.3 & 7.6 & 16.9 & 34.2 \\
RWKV & 1.75M & $\mathcal{O}(L)$ & 2.5 & 2.5 & 3.4 & 7.5 & 17.2 & 37.9 & 76.0 \\
RetNet & 2.38M & $\mathcal{O}(L)$ & 2.8 & 2.8 & 2.9 & 5.8 & 12.1 & 24.7 & 49.0 \\
xLSTM & 1.92M & $\mathcal{O}(L^2)$ & 3.6 & 11.9 & 41.0 & 151.9 & --- & --- & --- \\
Mamba & 2.01M & $\mathcal{O}(L)$ & 1.3 & 2.3 & 4.3 & 8.6 & 18.5 & 39.0 & 78.9 \\
Mamba-2 & 2.00M & $\mathcal{O}(L)$ & 1.8 & 1.8 & 2.3 & 4.9 & 10.5 & 25.6 & 52.9 \\
Naju (ours, $N\,64$) & 4.23M & $\mathcal{O}(L)$ & 2.9 & 5.3 & 10.4 & 21.5 & 44.0 & 87.1 & 173.3 \\
Naju (ours, $d\,128$, $N\,64$) & 1.18M & $\mathcal{O}(L)$ & 1.7 & 2.6 & 4.7 & 9.4 & 19.7 & 39.9 & 80.0 \\
\bottomrule
\end{tabular}\\[7pt]
{\bfseries (b) Peak memory (MB)}\\[2pt]
\begin{tabular}{lrc ccccccc}
\toprule
\textbf{Model} & \textbf{Params} & \textbf{Scaling} & \textbf{512} & \textbf{1024} & \textbf{2048} & \textbf{4096} & \textbf{8192} & \textbf{16k} & \textbf{32k} \\
\midrule
Transformer (math attn.)  & 11.81M & $\mathcal{O}(L^2)$ & 173 & 400 & 2{,}329 & 8{,}873 & \textbf{\textsc{oom}} & \textbf{\textsc{oom}} & \textbf{\textsc{oom}} \\
Transformer (flash attn.) & 11.81M & $\mathcal{O}(L^2)$ & 144 & 207 & 333 & 584 & 1{,}088 & 2{,}095 & 4{,}110 \\
HGRN & 1.32M & $\mathcal{O}(L)$ & 68 & 98 & 156 & 274 & 509 & 1{,}248 & 2{,}189 \\
GLA & 1.34M & $\mathcal{O}(L)$ & 70 & 102 & 165 & 291 & 544 & 1{,}317 & 2{,}327 \\
RWKV & 1.75M & $\mathcal{O}(L)$ & 106 & 171 & 301 & 561 & 1{,}081 & 2{,}126 & 4{,}203 \\
RetNet & 2.38M & $\mathcal{O}(L)$ & 94 & 145 & 247 & 451 & 858 & 1{,}936 & 3{,}560 \\
xLSTM & 1.92M & $\mathcal{O}(L^2)$ & 529 & 1{,}554 & 5{,}417 & 20{,}384 & \textbf{\textsc{oom}} & \textbf{\textsc{oom}} & \textbf{\textsc{oom}} \\
Mamba & 2.01M & $\mathcal{O}(L)$ & 119 & 196 & 350 & 657 & 1{,}273 & 2{,}505 & 4{,}968 \\
Mamba-2 & 2.00M & $\mathcal{O}(L)$ & 104 & 166 & 289 & 536 & 1{,}029 & 2{,}015 & 3{,}988 \\
Naju (ours, $N\,64$) & 4.23M & $\mathcal{O}(L)$ & 145 & 239 & 428 & 806 & 1{,}561 & 3{,}071 & 6{,}092 \\
Naju (ours, $d\,128$, $N\,64$) & 1.18M & $\mathcal{O}(L)$ & 87 & 135 & 231 & 424 & 811 & 1{,}583 & 3{,}127 \\
\bottomrule
\end{tabular}
\caption{\textbf{Forward latency and peak GPU memory versus sequence length.} All matched-width rows use $4$ layers, batch size $8$, and $d_{\text{model}}=256$; the additional $d=128$ Naju row reports its diagnostic-task configuration. Measurements are collected on one otherwise idle RTX~5090. Each latency cell reports the mean of $20$ forward iterations after $5$ warm-up iterations. \emph{Scaling} denotes the sequence-length complexity of the evaluated parallel forward implementation. Dashes in (a) indicate configurations that do not fit in memory, reported as \textsc{oom} in (b). Transformer parameter counts include the learned position table used in this benchmark.}
\label{tab:throughput}
\end{table}

\FloatBarrier
\section{Sensitivity and Ablation Studies}
\label{sec:ablations}
Because retention in Naju is controlled by the per-channel forget gate
$f_n=\sigma(a_n)$ with a learnable bias, its initial memory behavior depends
strongly on the two gate-bias initializations and on the dimension and scaling
of the recurrent state. This section studies those choices.
The experiments vary one primary design choice at a time, with selected
state-size comparisons included to test whether the observed trends persist
at the enlarged default state (all runs use Naju at
$d_{\text{model}}{=}128$, train at length $512$, and report length-extrapolated
accuracy; except for the knob under study, the base fixes $b_f{=}5$, $b_i{=}{-}2$,
$d_{\text{state}}{=}64$ with the parameter-free $1/\sqrt{d_{\text{state}}}$ readout
normalization of Eq.~\eqref{eq:readout_norm} and the diagonal feedthrough $D$ initialized at $0.01$). The
summary is twofold: the
preserve-first decoupled gate is robust, with each bias playing a distinct role
(Section~\ref{sec:bias_sweeps}); and the readout normalization substantially improves the reliability and worst-seed performance of the enlarged default state (Section~\ref{sec:norm_state}).

\subsection{Sensitivity to Gate-Bias Initialization}
\label{sec:bias_sweeps}

\paragraph{Forget-bias $b_f$ sweep at fixed $b_i$.}
We systematically sweep $b_f\in\{4,5,6\}$ at fixed $b_i=-2$ and
$d_{\text{state}}\in\{32,64\}$ over five seeds using the final default base,
including normalized readout and feedthrough $D$ initialized to $0.01$
(Table~\ref{tab:bf_sweep}).

At $d_{\text{state}}=64$, $b_f=5$ provides the best overall balance of mean
T4 extrapolation accuracy, worst-seed performance, and training reliability,
achieving $0.89\pm0.05$ at length $2048$, with all five seeds reaching the
training-length validation threshold. The smaller bias $b_f=4$ remains stable
at length $512$, but its mean T4 accuracy decreases to $0.80$ at length
$2048$. The larger bias $b_f=6$ exhibits substantially greater seed
sensitivity, with only three of five seeds reaching the validation threshold
and a worst-seed T4 accuracy of $0.15$.

This degradation coincides with the longer initial e-folding time constant of
$b_f=6$: Eq.~\eqref{eq:tau_e} gives approximately $404$ steps, compared with
approximately $149$ steps for $b_f=5$. This suggests that an excessively
near-lossless initialization can make early optimization less robust. At
$d_{\text{state}}=32$, all three biases train reliably and obtain similar mean
T4 performance, indicating that the sensitivity becomes more pronounced at
the enlarged state size. We therefore select $b_f=5$ as the default because
it provides the strongest and most reliable retention--overwrite balance at
$d_{\text{state}}=64$.

\begin{table}[t]
\centering\small
\begin{tabular}{ccccccc}
\toprule
$d_{\text{state}}$ & $b_f$ & \textbf{T2}@2048 & \textbf{T4}@512 & \textbf{T4}@2048 & \textbf{worst} & \textbf{successful seeds} \\
\midrule
$32$ & $4$ & $0.99${\scriptsize$\,\pm0.00$} & $0.98${\scriptsize$\,\pm0.01$} & $0.87${\scriptsize$\,\pm0.08$} & $0.75$ & $5/5$ \\
     & $5$ & $0.99${\scriptsize$\,\pm0.00$} & $0.98${\scriptsize$\,\pm0.01$} & $0.83${\scriptsize$\,\pm0.11$} & $0.70$ & $5/5$ \\
     & $6$ & $0.99${\scriptsize$\,\pm0.00$} & $0.99${\scriptsize$\,\pm0.00$} & $0.88${\scriptsize$\,\pm0.09$} & $0.72$ & $5/5$ \\
\midrule
$64$ & $4$ & $0.99${\scriptsize$\,\pm0.01$} & $0.99${\scriptsize$\,\pm0.01$} & $0.80${\scriptsize$\,\pm0.14$} & $0.55$ & $5/5$ \\
     & $5$ & $0.99${\scriptsize$\,\pm0.00$} & $0.99${\scriptsize$\,\pm0.00$} & $\mathbf{0.89}${\scriptsize$\,\pm0.05$} & $\mathbf{0.80}$ & $\mathbf{5/5}$ \\
     & $6$ & $0.99${\scriptsize$\,\pm0.01$} & $0.81${\scriptsize$\,\pm0.27$} & $0.60${\scriptsize$\,\pm0.27$} & $0.15$ & $3/5$ \\
\bottomrule
\end{tabular}
\caption{Forget-bias sweep at the default base ($b_i{=}{-}2$, normalized readout, feedthrough $D$ initialized at $0.01$), mean$\pm$std over $5$ seeds: retention (T2 kv\_spread at $2048$) and overwriting (T4 st\_hard at the training length $512$ and at $2048$). \textbf{worst} is the minimum seed on T4$@2048$; \textbf{successful seeds} counts runs whose validation accuracy at length
$512$ exceeds $0.9$.}
\label{tab:bf_sweep}
\end{table}

\paragraph{Input-bias $b_i$ sweep at fixed $b_f$.}
We isolate the input-gate bias by sweeping
$b_i\in\{-1,-2,-3,-4\}$ at fixed $b_f=5$ and
$d_{\text{state}}=64$ over five seeds using the default base, including
normalized readout and feedthrough $D$ initialized to $0.01$
(Table~\ref{tab:bi_sweep}).

Retention is largely insensitive to the input-bias initialization: T2 accuracy
remains between $0.99$ and $1.00$ across the sweep. In contrast, T4 exhibits
the best overall performance at $b_i=-2$, which achieves
$0.89\pm0.05$ at length $2048$ and a worst-seed accuracy of $0.80$.
The stronger initial write bias $b_i=-1$ produces greater training and
extrapolation variability, with only four of five seeds reaching the
training-length validation threshold. Weaker write biases remain reliable at
the training length but gradually reduce extrapolation performance, reaching
$0.86$ at $b_i=-3$ and $0.81$ at $b_i=-4$.

Together with the forget-bias sweep in Table~\ref{tab:bf_sweep}, these results
support distinct initialization roles for the two gates. The forget bias
primarily sets the initial retention time scale, whereas the input bias
controls the initial strength of new writes. We therefore use $b_i=-2$ as the
default because it provides the strongest and most reliable overwrite
performance while preserving near-perfect retention.
\begin{table}[t]
\centering\small
\begin{tabular}{ccccccc}
\toprule
$b_i$ & $\sigma(b_i)$ & \textbf{T2}@2048 & \textbf{T4}@512 & \textbf{T4}@2048 & \textbf{worst} & \textbf{successful seeds} \\
\midrule
$-1$ & $0.27$ & $1.00${\scriptsize$\,\pm0.00$} & $0.93${\scriptsize$\,\pm0.10$} & $0.78${\scriptsize$\,\pm0.17$} & $0.46$ & $4/5$ \\
$-2$ & $0.12$ & $0.99${\scriptsize$\,\pm0.00$} & $0.99${\scriptsize$\,\pm0.00$} & $\mathbf{0.89}${\scriptsize$\,\pm0.05$} & $\mathbf{0.80}$ & $\mathbf{5/5}$ \\
$-3$ & $0.05$ & $0.99${\scriptsize$\,\pm0.01$} & $0.98${\scriptsize$\,\pm0.02$} & $0.86${\scriptsize$\,\pm0.11$} & $0.73$ & $5/5$ \\
$-4$ & $0.02$ & $1.00${\scriptsize$\,\pm0.00$} & $0.99${\scriptsize$\,\pm0.01$} & $0.81${\scriptsize$\,\pm0.11$} & $0.66$ & $5/5$ \\
\bottomrule
\end{tabular}
\caption{Input-gate bias sweep at the default base ($b_f{=}5$, $d_{\text{state}}{=}64$, normalized readout, feedthrough $D$ initialized at $0.01$), mean$\pm$std over $5$ seeds: retention (T2 kv\_spread at $2048$) and overwriting (T4 st\_hard at the training length $512$ and at $2048$). \textbf{worst} is the minimum seed on T4$@2048$; \textbf{successful seeds} counts runs whose validation accuracy at length
$512$ exceeds $0.9$.}
\label{tab:bi_sweep}
\end{table}

\subsection{Readout Normalization versus State Size}
\label{sec:norm_state}

\begin{table}[t]
\centering
\small
\begin{tabular}{clccccc}
\toprule
$d_{\text{state}}$ &
\textbf{Readout gain} &
\textbf{T2}@2048 &
\textbf{T4}@512 &
\textbf{T4}@2048 &
\textbf{worst} &
\textbf{successful seeds} \\
\midrule
\multicolumn{7}{l}{
\emph{$b_f=5$ \quad
($\sigma(b_f)\approx0.993$, $\tau_e\approx149$)}
} \\
$32$
& $\alpha=1/\sqrt{d_{\text{state}}}$
& $0.99${\scriptsize$\,\pm0.00$}
& $0.98${\scriptsize$\,\pm0.01$}
& $0.83${\scriptsize$\,\pm0.11$}
& $0.70$
& $5/5$ \\
&
$\alpha=1$
& $0.99${\scriptsize$\,\pm0.00$}
& $0.99${\scriptsize$\,\pm0.01$}
& $0.88${\scriptsize$\,\pm0.11$}
& $0.67$
& $5/5$ \\
\rowcolor{gray!18}
$64$
& $\alpha=1/\sqrt{d_{\text{state}}}$ (default)
& $0.99${\scriptsize$\,\pm0.00$}
& $0.99${\scriptsize$\,\pm0.00$}
& $\mathbf{0.89}${\scriptsize$\,\pm0.05$}
& $\mathbf{0.80}$
& $\mathbf{5/5}$ \\
&
$\alpha=1$
& $0.99${\scriptsize$\,\pm0.00$}
& $0.95${\scriptsize$\,\pm0.09$}
& $0.80${\scriptsize$\,\pm0.17$}
& $0.48$
& $4/5$ \\
\midrule
\multicolumn{7}{l}{
\emph{$b_f=6$ \quad
($\sigma(b_f)\approx0.998$, $\tau_e\approx404$)}
} \\
$32$
& $\alpha=1/\sqrt{d_{\text{state}}}$
& $0.99${\scriptsize$\,\pm0.00$}
& $0.99${\scriptsize$\,\pm0.00$}
& $\mathbf{0.88}${\scriptsize$\,\pm0.09$}
& $0.72$
& $5/5$ \\
&
$\alpha=1$
& $1.00${\scriptsize$\,\pm0.00$}
& $0.99${\scriptsize$\,\pm0.00$}
& $0.87${\scriptsize$\,\pm0.10$}
& $0.68$
& $5/5$ \\
$64$
& $\alpha=1/\sqrt{d_{\text{state}}}$
& $0.99${\scriptsize$\,\pm0.01$}
& $0.81${\scriptsize$\,\pm0.27$}
& $0.60${\scriptsize$\,\pm0.27$}
& $0.15$
& $3/5$ \\
&
$\alpha=1$
& $0.98${\scriptsize$\,\pm0.01$}
& $0.92${\scriptsize$\,\pm0.09$}
& $0.73${\scriptsize$\,\pm0.19$}
& $0.49$
& $3/5$ \\
\bottomrule
\end{tabular}
\caption{\textbf{Readout gain versus state size.}
Results are mean$\pm$std over five seeds on T2 and T4, using $b_i=-2$ and
feedthrough $D$ initialized to $0.01$. \textbf{worst} is the minimum-seed
T4 accuracy at length $2048$, and \textbf{success} counts runs whose
validation accuracy at length $512$ exceeds $0.9$. Bold marks the highest T4
result within each forget-bias block; the shaded row is the default.}
\label{tab:norm_state}
\end{table}

\paragraph{Joint state-size and readout-gain comparison.}
We cross the state size
$d_{\text{state}}\in\{32,64\}$ with the dimension-corrected readout gain
$\alpha=1/\sqrt{d_{\text{state}}}$ and the unnormalized alternative
$\alpha=1$ at forget biases $b_f\in\{5,6\}$. All configurations use
$b_i=-2$, feedthrough $D$ initialized to $0.01$, and five random seeds
(Table~\ref{tab:norm_state}). Because T2 remains near saturated
($0.98$--$1.00$) throughout the sweep, the main discrimination comes from
T4, which tests overwriting under length extrapolation.

\paragraph{Normalization improves the enlarged state at $b_f=5$.}
At $d_{\text{state}}=32$, the two readout gains perform similarly:
$\alpha=1/\sqrt{d_{\text{state}}}$ obtains $0.83\pm0.11$ on T4 at length
$2048$, compared with $0.88\pm0.11$ for $\alpha=1$. At the enlarged
$d_{\text{state}}=64$ state, however, the dimension-corrected gain improves
both mean performance and seed robustness. It raises T4 accuracy from
$0.80\pm0.17$ to $0.89\pm0.05$, improves the worst seed from $0.48$ to
$0.80$, and increases the number of successful runs from $4/5$ to $5/5$.

This result is consistent with the readout-scale argument of
Section~\ref{sec:stability_consequence}: under the initialization assumptions
used there, the unnormalized state readout grows with
$\sqrt{d_{\text{state}}}$, whereas
$\alpha=1/\sqrt{d_{\text{state}}}$ compensates for this leading
dimension-dependent factor. The gain does not change the recurrent pole or
the BIBO stability result; it specifically rescales the state-readout pathway
relative to the direct and residual pathways.

\paragraph{Interaction with the forget bias.}
The normalization is a targeted correction rather than a universal
improvement. At $b_f=6$ and $d_{\text{state}}=32$, the two gains produce
nearly identical T4 results ($0.88\pm0.09$ versus $0.87\pm0.10$). At
$d_{\text{state}}=64$, neither gain yields reliable training: both
configurations have only $3/5$ successful seeds, and the unnormalized variant
obtains a higher mean T4 accuracy than the normalized one
($0.73\pm0.19$ versus $0.60\pm0.27$). Thus, readout rescaling alone does not
resolve the sensitivity associated with the more near-unit initialization
$b_f=6$, whose initial e-folding time constant is approximately $404$ steps,
compared with approximately $149$ steps for $b_f=5$.

Overall, the sweep identifies
$b_f=5$, $b_i=-2$, $d_{\text{state}}=64$, and
$\alpha=1/\sqrt{d_{\text{state}}}$ as the most reliable joint setting for the
contested retention and overwriting tasks. This configuration preserves
near-perfect T2 performance while obtaining the strongest mean and worst-seed
T4 results in the sweep. We therefore use it as the default in the main
experiments. The comparison with $\alpha=1$ should be interpreted as an
optimization-robustness ablation rather than a recurrent-stability ablation.

\section{Beyond the Diagnostic Suite}
\label{sec:beyond}

The diagnostic suite isolates two gate-level demands---retention and
overwriting---under controlled, sparsely supervised conditions. We next examine whether the resulting behavior extends beyond these synthetic
diagnostics to three external evaluations: multi-query associative recall
(MQAR), which probes associative-recall capacity and width--state scaling; the
Long Range Arena (LRA), which evaluates long-range sequence modeling across
multiple task domains; and WikiText-103 language modeling, which tests whether
the retain--write separation transfers to natural-language prediction under a
shared causal language-modeling protocol. 

\subsection{Multi-Query Associative Recall}
\label{sec:mqar}

\paragraph{Protocol.}
Multi-query associative recall (MQAR) is an established benchmark of
associative-recall capacity. A sequence presents key--value pairs from a
vocabulary of $8{,}192$ tokens, with each key written once, and the model must
return the corresponding value at subsequent query positions. In the taxonomy
of Section~\ref{sec:suite}, MQAR primarily evaluates retention: bindings are
not updated, and therefore the selective-overwriting demand of T4 is absent.
Unlike T2, however, MQAR provides dense supervision at multiple query
positions and explicitly varies the number of simultaneously stored bindings.

We use the canonical Zoology protocol: training mixes sequence lengths
$64$--$256$ and $4$--$64$ key--value pairs, whereas evaluation includes
extrapolation to length $1024$ with $256$ pairs. All models are evaluated in
the same two-layer harness. Each configuration is run with four learning
rates, and the best single-seed result is reported.
Across the standard Naju rows, the gate biases, readout normalization, and
feedthrough configuration remain fixed:
$b_f=5$, $b_i=-2$, $\alpha=1/\sqrt{d_{\text{state}}}$, and a learned
diagonal $D$ initialized to $0.01$. Model width and state size $N$ vary as
shown in Table~\ref{tab:mqar}.

\begin{table}[t]
\centering
\small
\setlength{\tabcolsep}{4pt}
\begin{tabular}{llcccccc}
\toprule
\textbf{Model} &
\textbf{$N$} &
\textbf{State/layer @$d\,256$} &
\textbf{$d\,64$} &
\textbf{$d\,128$} &
\textbf{$d\,256$} &
\textbf{$d\,512$} &
\textbf{kv$_{256}$@$d\,512$} \\
\midrule
Attention & --- & $512L$
& $1.000$ & $1.000$ & --- & --- & --- \\
xLSTM & --- & $65{,}536$
& $0.716$ & $0.898$ & $\mathbf{0.990}$ & $\mathbf{0.992}$
& $\mathbf{0.946}$ \\
\midrule
Mamba & $16$ & $8{,}192$
& $0.501$ & $0.707$ & $0.828$ & $\mathbf{0.879}$ & $0.345$ \\
Mamba-2 & $16$ & $8{,}192$
& $0.669$ & $0.816$ & $0.800$ & $0.837$ & $0.243$ \\
Naju (ours) & $16$ & $8{,}192$
& $0.649$ & $0.724$ & $\mathbf{0.871}$ & $0.877$ & $0.345$ \\
\quad $+$ optimizer-matched gate reparam.
& & &
$\mathbf{0.757}$ & $\mathbf{0.844}$ & $0.866$ & $0.877$
& $\mathbf{0.346}$ \\
\midrule
Mamba & $64$ & $32{,}768$
& $0.698$ & $0.633$ & $0.798$ & $0.828$ & $0.210$ \\
Mamba-2 & $64$ & $32{,}768$
& $\mathbf{0.853}$ & $\mathbf{0.896}$ & $0.909$ & $0.922$ & $0.523$ \\
Naju (ours) & $64$ & $32{,}768$
& $0.786$ & $0.831$ & $\mathbf{0.947}$ & $0.887$ & $0.384$ \\
\quad $+$ optimizer-matched gate reparam.
& & &
$0.790$ & $0.863$ & $0.936$ & $\mathbf{0.962}$
& $\mathbf{0.753}$ \\
\bottomrule
\end{tabular}
\caption{\textbf{MQAR accuracy under the canonical Zoology protocol.}
Each entry is the best result over the learning-rate sweep. Evaluation
includes extrapolation to length $1024$ with up to $256$ key--value pairs.
\emph{State/layer} reports recurrent-state elements at
$d_{\text{model}}=256$; attention instead uses a sequence-dependent KV cache
of $2Ld_{\text{model}}=512L$ elements. The final column reports the hardest
slice, with $256$ pairs at length $1024$. Indented rows use the
optimizer-matched gate reparameterization while leaving the Naju recurrence
unchanged. Bold marks the best recurrent result within each $N$ block and
column; attention is excluded from this comparison.}
\label{tab:mqar}
\end{table}

\paragraph{Results.}
Three observations emerge from Table~\ref{tab:mqar}.
First, Naju generally improves over the state-matched Mamba baseline. At
$N=16$, default Naju exceeds Mamba at widths $64$, $128$, and $256$, while
the two models are effectively tied at $d=512$ ($0.877$ versus $0.879$).
At $N=64$, default Naju exceeds Mamba at all four evaluated widths, with a
substantial advantage at $d=256$ ($0.947$ versus $0.798$).

Second, Mamba-2 is the stronger baseline at small widths. At $N=64$, it leads
default Naju at $d=64$ and $d=128$, whereas Naju overtakes it at $d=256$
($0.947$ versus $0.909$). At $d=512$, however, default Naju decreases to
$0.887$, suggesting that the gate parameterization becomes harder to optimize
as the model width increases.

Third, increasing the state size from $N=16$ to $N=64$ improves Naju at every
evaluated width. At the matched state budget of $32{,}768$ elements and
$d=256$, Naju achieves the strongest result among the three SSM models:
$0.947$, compared with $0.909$ for Mamba-2 and $0.798$ for Mamba.

The matrix-memory xLSTM remains the strongest recurrent model at the larger
widths, obtaining $0.990$ at $d=256$ and $0.992$ at $d=512$. Attention
achieves perfect accuracy in the reported small-width settings using a
sequence-dependent KV cache.

\paragraph{Optimizer-matched gate reparameterization.}
The decrease of default Naju from $0.947$ at $d=256$ to $0.887$ at $d=512$
motivates a width-aware treatment of the gate parameterization. Naju's
full-width gate projections read the $d_{\text{inner}}$-dimensional content
branch, so their optimization scale changes as the width increases.

In the reparameterized rows, the gate logits are divided by
$\sqrt{d_{\text{inner}}}$, while the corresponding weights are initialized
$\sqrt{d_{\text{inner}}}$ larger so that the initial function is unchanged.
To compensate for this parameterization change, the learning rate of the gate
parameter group is multiplied by $\sqrt{d_{\text{inner}}}$, whereas its
weight decay and Adam $\varepsilon$ are divided by
$\sqrt{d_{\text{inner}}}$. The architecture and recurrent equations remain
unchanged.

This correction has little effect where the default parameterization already
trains well, but substantially improves the widest $N=64$ model. At $d=512$,
accuracy rises from $0.887$ to $0.962$, and accuracy on the hardest
kv$_{256}$ slice rises from $0.384$ to $0.753$. The improvement indicates
that the default model's degradation at large width is primarily an
optimization-scaling issue rather than a limitation of the recurrence or
state capacity.

\paragraph{What MQAR does and does not show.}
MQAR evaluates retention and associative capacity, but it does not test
whether stale bindings can be selectively replaced. The benchmark is
therefore complementary to the diagnostic suite rather than a substitute for
it. For example, xLSTM performs strongly on MQAR but obtains lower T4
overwrite accuracy in Table~\ref{tab:summary}. Conversely, the relative
rankings of Naju and Mamba-2 change across MQAR, scattered long-range
retention, and recency-based overwriting.

Accordingly, the MQAR results support a narrower conclusion: Naju is
competitive in associative recall at a fixed recurrent-state budget, benefits
consistently from increasing $N$, and scales to the strongest SSM result at
$d=256$. The diagnostic suite supplies the separate evidence that this
retention capacity can coexist with selective overwriting.

\subsection{Long Range Arena}
\label{sec:lra}

\paragraph{Protocol.}
We evaluate the five core Long Range Arena tasks~\cite{tay2020lra} at their
full sequence lengths: ListOps ($L=2{,}000$), byte-level text classification
(Text, $L=4{,}000$), byte-level document matching
(Retrieval, two documents of $4{,}000$ bytes), sequential CIFAR
(Image, $L=1{,}024$), and Pathfinder ($L=1{,}024$).

The goal is a controlled comparison of sequence mixers rather than a
leaderboard claim. Naju, Mamba ($N=64$), Mamba-2 ($N=64$), and xLSTM use the
same six-layer bidirectional backbone with $d_{\text{model}}=128$, batch size
$32$, and AdamW with learning rate $3\times10^{-4}$ and $10\%$ warmup. Models
are trained for $50$ epochs, except Retrieval, which uses $30$ epochs. We use
no early stopping or data augmentation. Naju, Mamba, and Mamba-2 are evaluated
over three seeds; xLSTM uses one seed because its per-epoch wall-clock cost is
substantially higher.

The evaluated parallel xLSTM implementation materializes quadratic
intermediate tensors and exceeds the available GPU memory on ListOps, Text,
and Retrieval. It is therefore evaluated only on Image and Pathfinder, whose
sequence length is $1{,}024$. Naju uses block-level gradient checkpointing on
Retrieval; this changes memory use through recomputation but does not change
the forward computation. Published Transformer and S4 results are included
only as contextual reference points because they use different model
configurations and training recipes.

\begin{table*}[t]
\centering
\small
\setlength{\tabcolsep}{5pt}
\begin{tabular}{lcccccc}
\toprule
\textbf{Model}
& \textbf{ListOps}
& \textbf{Text}
& \textbf{Retrieval}
& \textbf{Image}
& \textbf{Pathf.}
& \textbf{Avg} \\
&
$2{,}000$
& $4{,}000$
& $2{\times}4{,}000$
& $1{,}024$
& $1{,}024$
& \\
\midrule
xLSTM
& \textsc{oom}
& \textsc{oom}
& \textsc{oom}
& $0.461$
& $\mathbf{0.738}$
& --- \\

Mamba ($N\,64$)
& $0.464{\scriptstyle\pm.010}$
& $0.885{\scriptstyle\pm.001}$
& $0.907{\scriptstyle\pm.003}$
& $0.513{\scriptstyle\pm.002}$
& $0.711{\scriptstyle\pm.032}$
& $0.696$ \\

Mamba-2 ($N\,64$)
& $0.473{\scriptstyle\pm.008}$
& $\mathbf{0.887}{\scriptstyle\pm.004}$
& $0.909{\scriptstyle\pm.001}$
& $0.468{\scriptstyle\pm.010}$
& $0.719{\scriptstyle\pm.004}$
& $0.691$ \\

Naju (ours)
& $\mathbf{0.554}{\scriptstyle\pm.012}$
& $0.868{\scriptstyle\pm.003}$
& $\mathbf{0.911}{\scriptstyle\pm.002}$
& $\mathbf{0.561}{\scriptstyle\pm.005}$
& $0.730{\scriptstyle\pm.028}$
& $\mathbf{0.725}$ \\
\midrule
\multicolumn{7}{l}{
\emph{Published reference results
(different configurations and training recipes)}
} \\
Transformer~\cite{tay2020lra}
& $0.364$
& $0.643$
& $0.575$
& $0.424$
& $0.714$
& $0.544$ \\

S4~\cite{gu2022s4}
& $0.596$
& $0.868$
& $0.909$
& $0.887$
& $0.942$
& $0.840$ \\
\bottomrule
\end{tabular}

\caption{\textbf{Full-length Long Range Arena results.}
Test accuracy is reported as mean$\pm$std over three seeds for Naju, Mamba,
and Mamba-2. xLSTM is evaluated with one seed because of its substantially
higher training cost. Bold marks the best model evaluated under our shared
training recipe for each task. The xLSTM \textsc{oom} entries indicate that
the evaluated parallel implementation does not fit in GPU memory at those
sequence lengths. \emph{Avg} is the unweighted mean over the five tasks and
is omitted for xLSTM because only two tasks complete. The bottom rows reproduce
published reference results obtained with different model configurations and
training recipes and are therefore not directly budget-matched baselines.}
\label{tab:lra}
\end{table*}

\paragraph{Results.}
Naju achieves the highest average accuracy among the models evaluated under
the shared training recipe, with an average of $0.725$, compared with $0.696$
for Mamba and $0.691$ for Mamba-2. The gains are concentrated on ListOps and
sequential CIFAR. On ListOps, Naju reaches $0.554$, compared with $0.464$ for
Mamba and $0.473$ for Mamba-2. On Image, it reaches $0.561$, compared with
$0.513$ and $0.468$, respectively.

The three scan-based models perform similarly on Text and Retrieval. Mamba-2
obtains the highest Text accuracy at $0.887$, while Naju achieves the highest
Retrieval accuracy at $0.911$; both differences are small. Pathfinder also
shows comparable performance across the recurrent models, with xLSTM,
Naju, Mamba-2, and Mamba obtaining $0.738$, $0.730$, $0.719$, and $0.711$,
respectively.

ListOps provides the clearest connection to the diagnostic suite. Solving
nested expressions requires a model to maintain intermediate results, update
them as subexpressions close, and prevent obsolete partial computations from
dominating the final representation. This is related to, but not identical
to, the stale-state suppression tested by T4. Naju's ListOps improvement is
therefore consistent with its strong state-update behavior in the diagnostic
suite, although LRA does not isolate retention and overwriting as cleanly as
the synthetic tasks.

\paragraph{Scope.}
The published S4 results are higher on several tasks, particularly Image and
Pathfinder, but they were obtained under a different model configuration and
training recipe. We therefore interpret Table~\ref{tab:lra} as evidence that
Naju remains competitive beyond the controlled diagnostic suite under one
shared experimental protocol, rather than as a state-of-the-art claim on the
LRA leaderboard.

\subsection{Subword Language Modeling on WikiText-103}
\label{sec:wt103}

\paragraph{Protocol.}
To test whether retain--write decoupling carries over from diagnostics and
long-range classification to real natural language, we train causal language
models on WikiText-103~\cite{merity2017pointer} using the raw corpus with
GPT-2 BPE~\cite{radford2019language}, a vocabulary of $50{,}257$, and one
end-of-text token per article. The resulting training set contains
$117.9$M tokens.

All models share the same language-modeling shell: tied input and output
embeddings, six pre-norm residual blocks, and
$d_{\text{model}}=256$. They differ in the sequence mixer: Naju, Mamba,
Mamba-2, or a Transformer using rotary attention~\cite{su2024roformer} and a
$4\times$ feed-forward expansion. The recurrent mixers use no explicit
positional encoding, whereas the Transformer uses RoPE.

The primary experiments use context length $1024$. We additionally train
separate runs at context lengths $2048$ and $4096$ using the same model
configurations and token budget. Training is matched by tokens rather than
optimizer steps: every run processes $32{,}768$ target tokens per update and
$1.2$B tokens in total using AdamW, BF16, a cosine schedule, and $5\%$
linear warmup. For each model, the learning rate is selected by validation
perplexity from a shared four-point grid using a $250$M-token pilot
(Appendix~\ref{app:wt103_lr}). The selected $L=1024$ learning rate is reused
without retuning for the $L=2048$ and $L=4096$ experiments. Checkpoints are
selected using validation perplexity, and the test set is evaluated once
after model selection. We report mean$\pm$std over three seeds; models with
the same seed receive identical batch sequences.

\begin{table}[htbp]
\centering\small
\setlength{\tabcolsep}{5pt}
\begin{tabular}{clcccc}
\toprule
$L$ & \textbf{Model} & \textbf{Params (mixer)} & \textbf{LR} & \textbf{Val PPL}~$\downarrow$ & \textbf{Test PPL}~$\downarrow$ \\
\midrule
$1024$ & Transformer (RoPE) & $17.6$M ($4.7$M) & $2{\times}10^{-3}$ & $26.45{\scriptstyle\pm.12}$ & $26.53{\scriptstyle\pm.12}$ \\
 & Mamba ($N\,16$)    & $15.5$M ($2.6$M) & $4{\times}10^{-3}$ & $28.28{\scriptstyle\pm.12}$ & $28.38{\scriptstyle\pm.10}$ \\
 & Mamba ($N\,64$)    & $15.9$M ($3.1$M) & $4{\times}10^{-3}$ & $27.23{\scriptstyle\pm.18}$ & $27.37{\scriptstyle\pm.15}$ \\
 & Mamba-2 ($N\,64$)  & $15.5$M ($2.6$M) & $4{\times}10^{-3}$ & $28.15{\scriptstyle\pm.14}$ & $28.31{\scriptstyle\pm.17}$ \\
 & Naju (ours)        & $18.8$M ($6.0$M) & $4{\times}10^{-3}$ & $\mathbf{26.06}{\scriptstyle\pm.10}$ & $\mathbf{26.20}{\scriptstyle\pm.17}$ \\
\midrule
$2048$ & Transformer (RoPE) & $17.6$M ($4.7$M) & $2{\times}10^{-3}$ & $25.72{\scriptstyle\pm.04}$ & $25.69{\scriptstyle\pm.06}$ \\
 & Mamba ($N\,16$)    & $15.5$M ($2.6$M) & $4{\times}10^{-3}$ & $27.67{\scriptstyle\pm.06}$ & $27.79{\scriptstyle\pm.12}$ \\
 & Mamba ($N\,64$)    & $15.9$M ($3.1$M) & $4{\times}10^{-3}$ & $26.50{\scriptstyle\pm.04}$ & $26.61{\scriptstyle\pm.06}$ \\
 & Mamba-2 ($N\,64$)  & $15.5$M ($2.6$M) & $4{\times}10^{-3}$ & $27.46{\scriptstyle\pm.10}$ & $27.56{\scriptstyle\pm.10}$ \\
 & Naju (ours)        & $18.8$M ($6.0$M) & $4{\times}10^{-3}$ & $\mathbf{25.38}{\scriptstyle\pm.20}$ & $\mathbf{25.53}{\scriptstyle\pm.19}$ \\
\midrule
$4096$ & Transformer (RoPE) & $17.6$M ($4.7$M) & $2{\times}10^{-3}$ & $25.34{\scriptstyle\pm.12}$ & $25.40{\scriptstyle\pm.11}$ \\
 & Mamba ($N\,16$)    & $15.5$M ($2.6$M) & $4{\times}10^{-3}$ & $27.36{\scriptstyle\pm.07}$ & $27.55{\scriptstyle\pm.05}$ \\
 & Mamba ($N\,64$)    & $15.9$M ($3.1$M) & $4{\times}10^{-3}$ & $26.17{\scriptstyle\pm.24}$ & $26.29{\scriptstyle\pm.25}$ \\
 & Mamba-2 ($N\,64$)  & $15.5$M ($2.6$M) & $4{\times}10^{-3}$ & $27.16{\scriptstyle\pm.05}$ & $27.23{\scriptstyle\pm.12}$ \\
 & Naju (ours)        & $18.8$M ($6.0$M) & $4{\times}10^{-3}$ & $\mathbf{25.08}{\scriptstyle\pm.05}$ & $\mathbf{25.18}{\scriptstyle\pm.07}$ \\
\bottomrule
\end{tabular}
\caption{WikiText-103 language modeling (raw corpus, GPT-2 BPE): validation
and test perplexity at a matched budget of $1.2$B training tokens
(mean$\pm$std over three seeds). All models share width, depth, context,
tokens per update, and the learning-rate candidate grid
(Appendix~\ref{app:wt103_lr}); parameter counts differ because the mixer
parameterizations differ and are reported as total (mixer-only). The
$L{=}2048$ and $L{=}4096$ blocks train at longer context with the same
tokens per update and the same $1.2$B budget; each model keeps its
$L{=}1024$ learning rate.}
\label{tab:wt103}
\end{table}

\paragraph{Results.}
Naju obtains the lowest validation and test perplexity among the evaluated
models at all three context lengths. At $L=1024$, its test perplexity is
$26.20\pm0.17$, compared with $26.53\pm0.12$ for the Transformer,
$27.37\pm0.15$ for state-matched Mamba with $N=64$, and
$28.31\pm0.17$ for Mamba-2. The corresponding margins are $0.33$, $1.17$,
and $2.11$ PPL, respectively, and are larger than the reported seed-level
standard deviations.

Increasing Mamba's state size from $N=16$ to $N=64$ improves its test
perplexity from $28.38$ to $27.37$, but a $1.17$-PPL gap to Naju remains at
the same state size. Mamba-2 provides the highest throughput among these
state-space baselines but obtains higher perplexity than state-matched Mamba
in this experiment. These comparisons indicate that Naju's result is not
explained by recurrent-state size alone. They should not, however, be read as
a parameter-matched comparison: Naju contains $18.8$M total parameters,
compared with $17.6$M for the Transformer and $15.5$--$15.9$M for the
Mamba variants.

Longer-context training improves every model while preserving the overall
ordering. At $L=2048$, Naju reaches a test perplexity of $25.53$, compared
with $25.69$ for the Transformer and $26.61$ for the strongest Mamba
configuration. At $L=4096$, Naju obtains $25.18$, compared with $25.40$ and
$26.29$, respectively. The Naju--Transformer margin therefore changes from
$0.33$ PPL at $L=1024$ to $0.16$ at $L=2048$ and $0.22$ at $L=4096$,
while the margin over state-matched Mamba remains approximately $1.1$ PPL.

Overall, the WikiText-103 results show that the benefit of the Naju recurrence
extends beyond controlled recall and state-tracking tasks. At matched width,
training-token budget, and language-modeling shell, Naju provides the lowest
perplexity among the evaluated sequence mixers at this model and data scale,
although it also uses a larger mixer parameterization.

\begin{table}[htbp]
\centering\small
\setlength{\tabcolsep}{5pt}
\begin{tabular}{clcccc}
\toprule
$d$ & \textbf{Model} & \textbf{Params (mixer)} & \textbf{LR} & \textbf{Val PPL}~$\downarrow$ & \textbf{Test PPL}~$\downarrow$ \\
\midrule
$256$ & Transformer (RoPE) & $17.6$M ($4.7$M) & $2{\times}10^{-3}$ & $26.45{\scriptstyle\pm.12}$ & $26.53{\scriptstyle\pm.12}$ \\
 & Mamba ($N\,64$)    & $15.9$M ($3.1$M) & $4{\times}10^{-3}$ & $27.23{\scriptstyle\pm.18}$ & $27.37{\scriptstyle\pm.15}$ \\
 & Mamba-2 ($N\,64$)  & $15.5$M ($2.6$M) & $4{\times}10^{-3}$ & $28.15{\scriptstyle\pm.14}$ & $28.31{\scriptstyle\pm.17}$ \\
 & Naju (ours)        & $18.8$M ($6.0$M) & $4{\times}10^{-3}$ & $\mathbf{26.06}{\scriptstyle\pm.10}$ & $\mathbf{26.20}{\scriptstyle\pm.17}$ \\
\midrule
$512$ & Transformer (RoPE) & $44.6$M ($18.9$M) & $2{\times}10^{-3}$ & $\mathbf{20.15}{\scriptstyle\pm.05}$ & $\mathbf{20.41}{\scriptstyle\pm.01}$ \\
 & Mamba ($N\,64$)    & $36.8$M ($11.1$M) & $4{\times}10^{-3}$ & $20.71{\scriptstyle\pm.12}$ & $21.00{\scriptstyle\pm.11}$ \\
 & Mamba-2 ($N\,64$)  & $35.7$M ($9.9$M) & $4{\times}10^{-3}$ & $21.29{\scriptstyle\pm.06}$ & $21.62{\scriptstyle\pm.13}$ \\
 & Naju (ours)        & $48.6$M ($22.9$M) & $4{\times}10^{-3}$ & $20.21{\scriptstyle\pm.04}$ & $20.49{\scriptstyle\pm.07}$ \\
\midrule
$1024$ & Transformer (RoPE) & $127.0$M ($75.5$M) & $2{\times}10^{-3}$ & $\mathbf{17.12}{\scriptstyle\pm.08}$ & $\mathbf{17.50}{\scriptstyle\pm.09}$ \\
 & Mamba ($N\,64$)    & $93.2$M ($41.8$M) & $4{\times}10^{-3}$ & $21.24{\scriptstyle\pm.66}$ & $21.69{\scriptstyle\pm.61}$ \\
 & Mamba-2 ($N\,64$)  & $90.3$M ($38.8$M) & $4{\times}10^{-3}$ & $17.99{\scriptstyle\pm.10}$ & $18.34{\scriptstyle\pm.09}$ \\
 & Naju (ours)        & $141.3$M ($89.9$M) & $4{\times}10^{-3}$ & $17.45{\scriptstyle\pm.03}$ & $17.82{\scriptstyle\pm.07}$ \\
\bottomrule
\end{tabular}
\caption{\textbf{WikiText-103 width scaling at fixed context.}
Validation and test perplexity are reported at context length $L=1024$ and a
matched budget of $1.2$B training tokens (mean$\pm$std over three seeds).
The $d=256$ block repeats the corresponding results from
Table~\ref{tab:wt103}. For $d\in\{512,1024\}$, all models are widened while
depth, context length, target tokens per update ($32{,}768$), and total
training tokens remain fixed. Each model reuses the learning rate selected at
$d=256$, so the wider results measure scaling under a fixed optimization
recipe rather than per-width retuning. Parameter counts are reported as total
parameters, with mixer-only parameters in parentheses. The $N=16$ Mamba
configuration is omitted to focus on the state-matched $N=64$ comparison; its
$d=256$ result is reported in Table~\ref{tab:wt103}.}
\label{tab:wt103_width}
\end{table}

\paragraph{Width scaling.}
Table~\ref{tab:wt103_width} examines width scaling from $d=256$ to $512$
and $1024$ at fixed context length $L=1024$, state size $N=64$, and a matched
training budget of $1.2$B tokens. The wider models reuse the learning rate
selected at $d=256$, so the results characterize scaling under a fixed
optimization recipe rather than independently tuned performance at each
width.

At $d=256$, Naju obtains the lowest test perplexity, reaching $26.20$
compared with $26.53$ for the Transformer, $27.37$ for Mamba, and $28.31$
for Mamba-2. At $d=512$, Naju and the Transformer become nearly tied
($20.49$ versus $20.41$), while both remain ahead of Mamba ($21.00$) and
Mamba-2 ($21.62$). At $d=1024$, the Transformer achieves the best result
($17.50$), followed by Naju ($17.82$) and Mamba-2 ($18.34$).
Thus, Naju's advantage at $d=256$ does not widen as model width increases:
the Transformer benefits more strongly from width scaling in this experiment,
Naju remains the strongest recurrent model at all three evaluated widths, although the Transformer becomes best overall at d=1024. 
Mamba-2 also improves consistently with
width, whereas Mamba improves at $d=512$ but degrades at $d=1024$ with
substantially larger seed variation.

This table fixes the recurrent state size at $N=64$ and therefore isolates
the width axis only. It does not determine whether increasing $N$ could
provide additional gains for Naju or the Mamba variants. In particular, the
relative ordering at larger widths may depend jointly on model width, state
size, and width-specific optimization, which motivates treating width and
state scaling as complementary rather than interchangeable sources of
capacity.

Finally, the comparison is matched in width, context, and training-token
budget, but not in parameter count. Naju has the largest mixer
parameterization at each width, whereas Mamba and Mamba-2 use fewer total and
mixer parameters.

\paragraph{Training throughput.} The quality gains in Table~\ref{tab:wt103} come with a higher training cost. Table~\ref{tab:wt103_eff} reports training throughput and peak memory under the corresponding WikiText-103 configurations. Measurements are collected on one otherwise idle RTX~5090 after $50$ warm-up steps; each reported value is the mean over the following $200$ training steps, with timing variation across measurement windows below $0.2\%$. 
The Naju measurements in Table~\ref{tab:wt103_eff} use the fused
chunk-parallel backend described in Appendix~\ref{app:cuda_scan}, for which
only the carry between chunk boundaries remains sequential. The remaining
efficiency gap therefore reflects inter-chunk carry propagation, recurrent
readout overhead, and incomplete fusion with the surrounding mixer
operations.

At $L=1024$, increasing Mamba's state size from $N=16$ to $N=64$ reduces throughput from $389.3$K to $315.3$K tokens/s. At the matched state size $N=64$, eager-mode Naju reaches $258.3$K tokens/s, corresponding to $0.82\times$ the throughput of Mamba and $0.56\times$ that of Mamba-2. Mamba-2 remains the fastest recurrent mixer throughout the evaluated context lengths.

Compilation recovers a substantial portion of Naju's eager-mode overhead. At $L=1024$, \texttt{torch.compile} raises Naju throughput from $258.3$K to $321.3$K tokens/s, an improvement of approximately $24\%$, while state-matched Mamba improves from $315.3$K to $348.6$K tokens/s, approximately $11\%$. The Mamba--Naju throughput ratio therefore narrows from $1.22\times$ in eager mode to $1.08\times$ after compilation.

Compiled Naju remains nearly constant as context length increases, reaching $318.4$K and $314.8$K tokens/s at $L=2048$ and $4096$, respectively. Compiled Mamba reaches $321.7$K and $312.3$K tokens/s at the same lengths. Thus, at $L=4096$, compiled Naju matches and slightly exceeds the state-matched Mamba implementation, although the difference is small. These results indicate that a meaningful part of Naju's eager-mode cost comes from implementation and kernel-fusion overhead rather than from its asymptotic sequence complexity.

\begin{table}[htbp]
\centering\small
\begin{tabular}{clccc}
\toprule
$L$ & \textbf{Model} & \textbf{Train tokens/s} & \textbf{ms\,/\,micro-step} & \textbf{Peak mem (MB)} \\
\midrule
$1024$ & Transformer (math attn.)  & $242.5$K & $33.8$ & $7{,}818$ \\
 & Transformer (flash attn.) & $411.6$K & $19.9$ & $6{,}004$ \\
 & Mamba ($N\,16$)           & $389.3$K & $21.0$ & $5{,}912$ \\
 & Mamba ($N\,64$)           & $315.3$K & $26.0$ & $5{,}950$ \\
 & \quad $+$ torch.compile   & $348.6$K & $23.5$ & $5{,}891$ \\
 & Mamba-2 ($N\,64$)         & $465.6$K & $17.6$ & $5{,}619$ \\
 & \quad $+$ torch.compile   & $479.9$K & $17.1$ & $5{,}558$ \\
 & Naju (ours, $N\,64$) & $258.3$K & $31.7$ & $6{,}276$ \\
 & \quad $+$ torch.compile   & $321.3$K & $25.5$ & $6{,}177$ \\
\midrule
$2048$ & Transformer (math attn.)  & $168.8$K & $48.5$ & $9{,}627$ \\
 & Transformer (flash attn.) & $396.4$K & $20.7$ & $6{,}004$ \\
 & Mamba ($N\,16$)           & $379.9$K & $21.6$ & $5{,}909$ \\
 & Mamba ($N\,64$)           & $293.7$K & $27.9$ & $5{,}941$ \\
 & \quad $+$ torch.compile   & $321.7$K & $25.5$ & $5{,}884$ \\
 & Mamba-2 ($N\,64$)         & $466.1$K & $17.6$ & $5{,}610$ \\
 & \quad $+$ torch.compile   & $479.2$K & $17.1$ & $5{,}557$ \\
 & Naju (ours, $N\,64$) & $258.5$K & $31.7$ & $6{,}276$ \\
 & \quad $+$ torch.compile   & $318.4$K & $25.7$ & $6{,}177$ \\
\midrule
$4096$ & Transformer (math attn.)  & $102.7$K & $79.8$ & $13{,}254$ \\
 & Transformer (flash attn.) & $366.6$K & $22.3$ & $6{,}007$ \\
 & Mamba ($N\,16$)           & $374.2$K & $21.9$ & $5{,}909$ \\
 & Mamba ($N\,64$)           & $286.0$K & $28.6$ & $5{,}941$ \\
 & \quad $+$ torch.compile   & $312.3$K & $26.2$ & $5{,}884$ \\
 & Mamba-2 ($N\,64$)         & $463.6$K & $17.7$ & $5{,}610$ \\
 & \quad $+$ torch.compile   & $477.8$K & $17.1$ & $5{,}557$ \\
 & Naju (ours, $N\,64$) & $257.2$K & $31.8$ & $6{,}276$ \\
 & \quad $+$ torch.compile   & $\mathbf{314.8}$K & $26.0$ & $6{,}177$ \\
\bottomrule
\end{tabular}
\caption{\textbf{WikiText-103 training efficiency.} All measurements use BF16 on one otherwise idle RTX~5090. Micro-batch sizes are $8$, $4$, and $2$ at $L=1024$, $2048$, and $4096$, respectively, so each micro-step contains the same number of target tokens; AdamW updates are performed after four micro-steps. Each cell is the mean over $200$ measured training steps following $50$ warm-up steps. Indented rows use \texttt{torch.compile} with max-autotune. Bold highlights the long-context point at which compiled Naju slightly exceeds compiled, state-matched Mamba; it does not indicate the fastest model overall.}
\label{tab:wt103_eff}
\end{table}

\paragraph{Gate-rank ablation.}
Naju has linear sequence-length scaling, but its eager-mode implementation
has a larger execution constant than the optimized Mamba kernels. To
distinguish the cost of the gate projections from that of recurrent
execution, we factorize the two full-width gate projections to ranks $64$
and $32$ (Table~\ref{tab:gate_rank}). The rank-$32$ variant removes
approximately $87\%$ of the gate-projection FLOPs and reduces the mixer-only
parameter count from $6.0$M to $3.2$M. Nevertheless, training throughput
changes by less than $1\%$, while forward latency improves by at most
approximately $10\%$ over sequence lengths $2{,}048$--$32{,}768$.

These results indicate that the gate projections are not the dominant runtime
cost at the evaluated width $d_{\text{model}}=256$. Their capacity is
nonetheless important: the rank-$64$ and rank-$32$ variants increase test
perplexity by $0.6$ and $0.8$, respectively. Thus, the full-width retain and
write gates perform quality-bearing computation that cannot be removed
without degrading language-modeling accuracy. Because gate-projection cost
grows quadratically with width, however, this runtime conclusion is specific
to the evaluated configuration.

\begin{table}[t]
\centering\small
\setlength{\tabcolsep}{3.5pt}
\begin{tabular}{lccccccc}
\toprule
\textbf{Model} & \textbf{Params (mixer)} & \textbf{LR} & \textbf{Val PPL}~$\downarrow$ & \textbf{Test PPL}~$\downarrow$ & \textbf{Tok/s} & \textbf{ms/$\mu$step} & \textbf{Mem (MB)} \\
\midrule
Naju (full-rank)   & $18.8$M ($6.0$M) & $4{\times}10^{-3}$ & $\mathbf{26.06}{\scriptstyle\pm.10}$ & $\mathbf{26.20}{\scriptstyle\pm.17}$ & $258.3$K & $31.7$ & $6{,}276$ \\
Naju (rank-$64$)   & $16.5$M ($3.6$M) & $4{\times}10^{-3}$ & $26.63{\scriptstyle\pm.08}$ & $26.80{\scriptstyle\pm.13}$ & $260.8$K & $31.4$ & $6{,}246$ \\
Naju (rank-$32$)   & $16.1$M ($3.2$M) & $4{\times}10^{-3}$ & $26.92{\scriptstyle\pm.11}$ & $27.00{\scriptstyle\pm.12}$ & $260.7$K & $31.4$ & $6{,}233$ \\
\bottomrule
\end{tabular}
\caption{\textbf{Gate-rank ablation on WikiText-103.}
Efficiency is measured at $L=1024$ under the protocol of
Table~\ref{tab:wt103_eff}. Factorizing the retain- and write-gate projections
to ranks $64$ and $32$ removes approximately $75\%$ and $87\%$ of their
projection FLOPs and reduces the mixer-only parameter count by $2.4$M and
$2.8$M, respectively. Throughput and memory change by less than $1\%$ at
$d_{\text{model}}=256$, while test perplexity increases by $0.6$ and $0.8$.
The runtime conclusion is specific to the evaluated width because
gate-projection cost grows quadratically with width.}
\label{tab:gate_rank}
\end{table}

\section{Conclusion}

We presented Naju, a native discrete selective state space model that
decouples retention from writing. In Naju, the learned forget gate directly
parameterizes the discrete recurrent pole, while an independent input gate
controls the write gain. This removes the structural constraint of
complementary single-gate updates, in which stronger writing necessarily
reduces retention. At four times the training length, Naju achieves $0.99$
accuracy on the retention-intensive T2 task and $0.89$ on the
overwrite-intensive T4 task, providing the strongest joint balance between
the two axes among the evaluated models.

The frozen-gate analysis gives the recurrence a direct first-order
discrete-filter interpretation: the forget gate determines the pole and the
input gate determines the write gain. Because the sigmoid-parameterized pole
lies in $(0,1)$, each frozen local system is Schur stable by construction.
Under uniformly bounded gate logits and bounded inputs and readouts, the
time-varying recurrence additionally satisfies a fading-memory and BIBO
bound without requiring an explicit stability regularizer.

The empirical benefits extend beyond the controlled diagnostic suite. At
$d_{\text{model}}=256$ and a matched $1.2$B-token training budget on
WikiText-103, Naju obtains the lowest perplexity among the evaluated
Transformer, Mamba, and Mamba-2 mixers across training contexts
$L\in\{1024,2048,4096\}$. The width-scaling experiments provide a more
qualified picture: Naju leads at $d=256$, remains close to the Transformer at
$d=512$, and is surpassed by the Transformer at $d=1024$. Thus, the observed
advantage does not increase uniformly with model scale, and the relative
ordering depends on context length, model width, state size, and optimization.
The comparison is also width- and token-budget-matched rather than
parameter-matched, because Naju uses a larger mixer parameterization.

On the systems side, Naju retains linear time and memory complexity in
sequence length. Its affine recurrence admits an exact chunk-parallel
decomposition in which intra-chunk computation is evaluated in parallel and
only the carry between chunks remains recurrent. The fused chunk backend
substantially reduces scan time while preserving pilot validation perplexity,
demonstrating implementation headroom without changing the recurrence itself.
The gate-rank ablation further shows that reducing the gate-projection FLOPs
has little effect on runtime at $d_{\text{model}}=256$ but degrades language
modeling quality. This indicates that the full-width retain and write gates
perform quality-bearing computation, while a substantial portion of the
measured execution overhead arises from the recurrent scan and its surrounding
operations rather than from gate arithmetic alone.

Naju still compresses the entire history into a fixed-size recurrent state and
therefore does not provide unbounded exact recall. Its contribution is instead
to make fixed-state memory more effective by supporting long retention and
selective overwriting within the same recurrence. For applications that also
require direct access to an unbounded set of past tokens, hybrid architectures
that combine recurrent and attention layers, such as
Jamba~\cite{lieber2024jamba} and Griffin~\cite{de2024griffin}, provide a
natural extension. Further directions include content-addressed erasure for
more precise far-distance overwriting, state- and width-aware optimization,
and lower-rank or structured gate parameterizations that improve efficiency at
larger widths while preserving the quality contribution of decoupled gating.

\bibliographystyle{plainnat}
\bibliography{references}

\appendix

\section{Proof of the Fading-Memory and BIBO Bounds}
\label{app:stability_bound}

We prove the bound for one expanded channel with a
$d_{\text{state}}$-dimensional state vector. Consider
\begin{equation}
x_n
=
f_n\odot x_{n-1}
+
v_n,
\qquad
f_n=\sigma(a_n),
\notag
\end{equation}
where
$x_n,f_n,a_n,v_n\in\mathbb{R}^{d_{\text{state}}}$.
Assume that the forget logits are uniformly upper-bounded coordinatewise:
there exists $A_{\max}<\infty$ such that
\begin{equation}
a_{n,k}\le A_{\max}
\qquad
\text{for all } n \text{ and } k.
\notag
\end{equation}
Then
\begin{equation}
0<f_{n,k}\le\rho:=\sigma(A_{\max})<1
\qquad
\text{for all } n \text{ and } k.
\notag
\end{equation}
Assume also that
\begin{equation}
\|v_n\|_2\le M_v
\qquad
\text{for all } n.
\notag
\end{equation}

Unrolling the recurrence gives
\begin{equation}
x_n
=
\left(
\bigodot_{j=1}^{n}f_j
\right)
\odot x_0
+
\sum_{m=1}^{n}
\left(
\bigodot_{j=m+1}^{n}f_j
\right)
\odot v_m,
\notag
\end{equation}
where the empty Hadamard product is the all-ones vector. Since every
coordinate of $f_j$ is at most $\rho$,
\begin{align}
\|x_n\|_2
&\le
\rho^n\|x_0\|_2
+
\sum_{m=1}^{n}
\rho^{n-m}\|v_m\|_2
\notag\\
&\le
\rho^n\|x_0\|_2
+
M_v\sum_{m=1}^{n}\rho^{n-m}
\notag\\
&\le
\rho^n\|x_0\|_2
+
\frac{M_v}{1-\rho}.
\notag
\end{align}
Thus, the state is uniformly bounded for every bounded initial state and
uniformly bounded write sequence.

Moreover, the contribution of the write at time $m$ to the state at time $n$
is
\begin{equation}
\left(
\bigodot_{j=m+1}^{n}f_j
\right)
\odot v_m,
\notag
\end{equation}
and satisfies
\begin{equation}
\left\|
\left(
\bigodot_{j=m+1}^{n}f_j
\right)
\odot v_m
\right\|_2
\le
\rho^{n-m}\|v_m\|_2.
\notag
\end{equation}
Hence, the influence of a past write decays geometrically with its temporal
distance. This establishes the fading-memory bound.

For the corresponding single-channel readout,
\begin{equation}
y_n
=
\alpha C_n^{\top}x_n
+
Dh_n,
\notag
\end{equation}
assume that
\begin{equation}
\|C_n\|_2\le M_C,
\qquad
|D|\le M_D,
\qquad
|h_n|\le M_h.
\notag
\end{equation}
Then, by the Cauchy--Schwarz inequality,
\begin{align}
|y_n|
&=
\left|
\alpha C_n^{\top}x_n
+
Dh_n
\right|
\notag\\
&\le
\alpha\|C_n\|_2\|x_n\|_2
+
|D|\,|h_n|
\notag\\
&\le
\alpha M_C
\left(
\rho^n\|x_0\|_2
+
\frac{M_v}{1-\rho}
\right)
+
M_D M_h.
\notag
\end{align}
Therefore, the output is uniformly bounded for every bounded initial state
and uniformly bounded write and readout sequences. In particular, for
$x_0=0$, every uniformly bounded input sequence produces a uniformly bounded
output sequence, establishing BIBO stability.

\section{Per-Task Results with Length Extrapolation}
\label{app:pertask}

This appendix expands the summary of Table~\ref{tab:summary} into full
per-task results. For each diagnostic task, we report accuracy at lengths
$512$, $1024$, and $2048$ as mean$\pm$std over five seeds. The
organization proceeds from associative recall (T1 and T2) to current-state
tracking (T3 and T4).

\subsection{Associative Recall}
\label{sec:results}

\begin{table}[htbp]
\centering
\small
\begin{tabular}{lcccc}
\toprule
\textbf{Model} & \textbf{Params} & \textbf{512} & \textbf{1024} & \textbf{2048} \\
\midrule
Transformer  & 4.29M & $0.13${\scriptsize$\,\pm0.01$} & $0.12${\scriptsize$\,\pm0.00$} & $0.11${\scriptsize$\,\pm0.01$} \\
\;{\scriptsize$\hookrightarrow$ train\,/\,best-val @512} & & {\scriptsize$0.72\,/\,0.13$} & & \\
Transformer ($150$ ep.) & 4.29M & $0.13${\scriptsize$\,\pm0.01$} & $0.11${\scriptsize$\,\pm0.02$} & $0.11${\scriptsize$\,\pm0.01$} \\
\;{\scriptsize$\hookrightarrow$ train\,/\,best-val @512} & & {\scriptsize$0.98\,/\,0.14$} & & \\
Mamba        & 1.84M & $1.00${\scriptsize$\,\pm0.00$} & $1.00${\scriptsize$\,\pm0.00$} & $1.00${\scriptsize$\,\pm0.00$} \\
Mamba-2      & 0.50M & $1.00${\scriptsize$\,\pm0.00$} & $1.00${\scriptsize$\,\pm0.00$} & $1.00${\scriptsize$\,\pm0.00$} \\
HGRN         & 0.29M & $0.11${\scriptsize$\,\pm0.01$} & $0.12${\scriptsize$\,\pm0.00$} & $0.11${\scriptsize$\,\pm0.01$} \\
GLA          & 0.31M & $1.00${\scriptsize$\,\pm0.00$} & $1.00${\scriptsize$\,\pm0.00$} & $1.00${\scriptsize$\,\pm0.00$} \\
RWKV         & 0.51M & $0.10${\scriptsize$\,\pm0.04$} & $0.11${\scriptsize$\,\pm0.03$} & $0.10${\scriptsize$\,\pm0.03$} \\
RetNet       & 0.56M & $1.00${\scriptsize$\,\pm0.00$} & $1.00${\scriptsize$\,\pm0.00$} & $1.00${\scriptsize$\,\pm0.00$} \\
xLSTM        & 0.46M & $1.00${\scriptsize$\,\pm0.00$} & $1.00${\scriptsize$\,\pm0.00$} & $1.00${\scriptsize$\,\pm0.00$} \\
Naju (ours)  & 1.09M & $1.00${\scriptsize$\,\pm0.00$} & $1.00${\scriptsize$\,\pm0.00$} & $1.00${\scriptsize$\,\pm0.00$} \\
\bottomrule
\end{tabular}
\caption{Key-value retrieval (T1) test accuracy, mean$\pm$std over $5$ seeds (train @ $512$, evaluated at $512/1024/2048$). Naju, Mamba, GLA, RetNet, and xLSTM solve it; Transformer, HGRN, and RWKV do not at this matched budget. The ($150$ ep.)\ row is an extended $150$-epoch budget ($5$ seeds; cf.\ Table~\ref{tab:state}). The extended budget does not rescue the Transformer here: train accuracy reaches ${\approx}1.0$ while test stays at chance---memorization, not slow convergence. Indented sub-rows: final-epoch train / best-validation accuracy at the training length.}
\label{tab:main}
\end{table}
Table~\ref{tab:main} reports key-value retrieval accuracy on T1. Naju,
Mamba, Mamba-2, GLA, RetNet, and xLSTM maintain near-perfect accuracy
through length $2048$, whereas the Transformer, HGRN, and RWKV remain near
chance under the matched training budget. T1 therefore confirms that several
recurrent mechanisms can solve the contiguous retrieval setting; the scattered
T2 task below is needed to distinguish their retention under substantially
longer query-to-fact distances.

\subsection{Long-Range Retention via Length Extrapolation}
\label{sec:lenextrap}

\begin{table}[htbp]
\centering
\small
\begin{tabular}{lccc}
\toprule
\textbf{Model (train @ 512)} & \textbf{512} & \textbf{1024} & \textbf{2048} \\
\midrule
Transformer  & $0.12${\scriptsize$\,\pm0.00$} & $0.13${\scriptsize$\,\pm0.01$} & $0.12${\scriptsize$\,\pm0.01$} \\
\;{\scriptsize$\hookrightarrow$ train\,/\,best-val @512} & {\scriptsize$0.99\,/\,0.13$} & & \\
Transformer ($150$ ep.) & $0.13${\scriptsize$\,\pm0.01$} & $0.11${\scriptsize$\,\pm0.02$} & $0.11${\scriptsize$\,\pm0.01$} \\
\;{\scriptsize$\hookrightarrow$ train\,/\,best-val @512} & {\scriptsize$1.00\,/\,0.13$} & & \\
Mamba        & $0.67${\scriptsize$\,\pm0.33$} & $0.67${\scriptsize$\,\pm0.33$} & $0.62${\scriptsize$\,\pm0.30$} \\
Mamba-2      & $0.63${\scriptsize$\,\pm0.35$} & $0.56${\scriptsize$\,\pm0.33$} & $0.33${\scriptsize$\,\pm0.17$} \\
HGRN         & $0.06${\scriptsize$\,\pm0.00$} & $0.07${\scriptsize$\,\pm0.00$} & $0.06${\scriptsize$\,\pm0.00$} \\
GLA          & $0.08${\scriptsize$\,\pm0.00$} & $0.08${\scriptsize$\,\pm0.00$} & $0.07${\scriptsize$\,\pm0.00$} \\
RWKV         & $0.06${\scriptsize$\,\pm0.00$} & $0.06${\scriptsize$\,\pm0.00$} & $0.06${\scriptsize$\,\pm0.00$} \\
RetNet       & $0.91${\scriptsize$\,\pm0.08$} & $0.51${\scriptsize$\,\pm0.06$} & $0.30${\scriptsize$\,\pm0.03$} \\
xLSTM        & $1.00${\scriptsize$\,\pm0.00$} & $1.00${\scriptsize$\,\pm0.00$} & $1.00${\scriptsize$\,\pm0.00$} \\
Naju (ours)  & $1.00${\scriptsize$\,\pm0.00$} & $1.00${\scriptsize$\,\pm0.00$} & $0.99${\scriptsize$\,\pm0.00$} \\
\bottomrule
\end{tabular}
\caption{Length extrapolation on the scattered key-value task T2 ($32$ facts spread across the sequence; chance $=1/16$), mean$\pm$std over $5$ seeds; trained at $512$ and tested longer without retraining. Only Naju (ours) and xLSTM stay flat at distance (both ${\approx}1.00$ at $2048$); Mamba/Mamba-2/RetNet solve the training length but degrade sharply under extrapolation ($0.62$/$0.33$/$0.30$), and GLA/RWKV/HGRN/Transformer are at chance. The ($150$ ep.)\ row is an extended $150$-epoch budget ($5$ seeds), which does not help---train accuracy ${\approx}1.0$ with test at chance (memorization). Indented sub-rows: final-epoch train / best-validation accuracy at the training length.}
\label{tab:extrap}
\end{table}
\paragraph{Scattered associative recall (T2).}
The contiguous task above saturates near the ceiling for several models, so it
does not clearly separate their long-range retention behavior. We therefore
use the scattered key-value task T2, in which $32$ facts are distributed
throughout the sequence with padding gaps and queried at the end. Models are
trained at length $512$ and evaluated without retraining at lengths
$512$, $1024$, and $2048$.

Table~\ref{tab:extrap} shows a clear separation. Naju and xLSTM remain near
perfect at length $2048$, whereas Mamba, Mamba-2, and RetNet degrade as the
query-to-fact distance increases. GLA, HGRN, RWKV, and the Transformer remain
near chance on T2. In particular, GLA solves the contiguous T1 task but fails
when the facts are scattered over longer distances. These results are
consistent with insufficient retention across the intervening gaps, although
the experiment does not by itself isolate the architecture-specific cause of
the degradation. Naju retains the scattered bindings with essentially no loss
through four times the training length.

\subsection{Current-State Tracking and Overwriting}
\label{sec:overwrite}

\begin{table}[htbp]
\centering
\small
\begin{tabular}{lccc}
\toprule
\textbf{Model (train @ 512)} & \textbf{512} & \textbf{1024} & \textbf{2048} \\
\midrule
Transformer  & $0.71${\scriptsize$\,\pm0.35$} & $0.07${\scriptsize$\,\pm0.00$} & $0.07${\scriptsize$\,\pm0.00$} \\
\;{\scriptsize$\hookrightarrow$ train\,/\,best-val @512} & {\scriptsize$0.73\,/\,0.72$} & & \\
Transformer ($150$ ep.) & $0.92${\scriptsize$\,\pm0.08$} & $0.07${\scriptsize$\,\pm0.01$} & $0.06${\scriptsize$\,\pm0.01$} \\
\;{\scriptsize$\hookrightarrow$ train\,/\,best-val @512} & {\scriptsize$0.93\,/\,0.92$} & & \\
Mamba        & $1.00${\scriptsize$\,\pm0.00$} & $1.00${\scriptsize$\,\pm0.00$} & $1.00${\scriptsize$\,\pm0.00$} \\
Mamba-2      & $1.00${\scriptsize$\,\pm0.00$} & $1.00${\scriptsize$\,\pm0.00$} & $0.99${\scriptsize$\,\pm0.01$} \\
HGRN         & $1.00${\scriptsize$\,\pm0.00$} & $1.00${\scriptsize$\,\pm0.00$} & $1.00${\scriptsize$\,\pm0.00$} \\
GLA          & $1.00${\scriptsize$\,\pm0.00$} & $1.00${\scriptsize$\,\pm0.00$} & $1.00${\scriptsize$\,\pm0.00$} \\
RWKV         & $0.97${\scriptsize$\,\pm0.05$} & $0.96${\scriptsize$\,\pm0.08$} & $0.95${\scriptsize$\,\pm0.10$} \\
RetNet       & $1.00${\scriptsize$\,\pm0.00$} & $0.92${\scriptsize$\,\pm0.02$} & $0.74${\scriptsize$\,\pm0.03$} \\
xLSTM        & $1.00${\scriptsize$\,\pm0.00$} & $1.00${\scriptsize$\,\pm0.00$} & $0.95${\scriptsize$\,\pm0.02$} \\
Naju (ours)  & $1.00${\scriptsize$\,\pm0.00$} & $1.00${\scriptsize$\,\pm0.00$} & $0.94${\scriptsize$\,\pm0.01$} \\
\bottomrule
\end{tabular}
\caption{Current-state tracking (T3) accuracy, mean$\pm$std over $5$ seeds (train @ $512$, evaluated at $512/1024/2048$). Most recurrent models solve it and hold under extrapolation; RetNet decays with distance. The Transformer is highly seed-variable within the matched $50$-epoch budget ($0.71\!\pm\!0.35$ at the training length: it groks on some seeds and stays near chance on others); with the extended $150$-epoch budget (the $150$ ep.\ row, $5$ seeds) it mostly groks the training length ($0.92\!\pm\!0.08$ at $512$), confirming slow convergence---but it still fails to extrapolate ($0.07/0.06$ at $1024/2048$), so the gap is a length-generalization failure, not merely an optimization one. Indented sub-rows: final-epoch train / best-validation accuracy at the training length---unlike T1/T2/T4, train and validation rise \emph{together} here.}
\label{tab:state}
\end{table}
\paragraph{Marked current-state tracking (T3).}
T3 evaluates whether a model can maintain the current value of an entity when
the answer-determining update is explicitly marked. Table~\ref{tab:state}
shows that most recurrent models solve this easier setting and extrapolate
well beyond the training length. Mamba, Mamba-2, HGRN, and GLA remain near
perfect through length $2048$, while RWKV, xLSTM, and Naju retain accuracies
of $0.95$, $0.95$, and $0.94$, respectively. RetNet degrades more
substantially, from $1.00$ at length $512$ to $0.74$ at length $2048$.

The Transformer shows a different failure mode. Under the matched
$50$-epoch budget, its training-length performance is highly seed-variable.
Extending training to $150$ epochs raises both training and validation
accuracy at length $512$, showing that the training-length task can
eventually be learned. However, its accuracy remains near chance at lengths
$1024$ and $2048$. The remaining failure is therefore one of length
extrapolation rather than merely slow optimization. Overall, T3 shows that
when the relevant update is explicitly identified, many recurrent
architectures can preserve the current value over long distances.

\begin{table}[htbp]
\centering
\small
\begin{tabular}{lccc}
\toprule
\textbf{Model (train @ 512)} & \textbf{512} & \textbf{1024} & \textbf{2048} \\
\midrule
Transformer  & $0.11${\scriptsize$\,\pm0.01$} & $0.06${\scriptsize$\,\pm0.00$} & $0.07${\scriptsize$\,\pm0.00$} \\
\;{\scriptsize$\hookrightarrow$ train\,/\,best-val @512} & {\scriptsize$0.71\,/\,0.12$} & & \\
Transformer ($150$ ep.) & $0.12${\scriptsize$\,\pm0.00$} & $0.07${\scriptsize$\,\pm0.00$} & $0.06${\scriptsize$\,\pm0.01$} \\
\;{\scriptsize$\hookrightarrow$ train\,/\,best-val @512} & {\scriptsize$1.00\,/\,0.13$} & & \\
Mamba        & $0.84${\scriptsize$\,\pm0.21$} & $0.82${\scriptsize$\,\pm0.25$} & $0.77${\scriptsize$\,\pm0.27$} \\
Mamba-2      & $0.91${\scriptsize$\,\pm0.08$} & $0.85${\scriptsize$\,\pm0.12$} & $0.68${\scriptsize$\,\pm0.15$} \\
HGRN         & $0.13${\scriptsize$\,\pm0.02$} & $0.11${\scriptsize$\,\pm0.02$} & $0.08${\scriptsize$\,\pm0.01$} \\
GLA          & $0.98${\scriptsize$\,\pm0.01$} & $0.96${\scriptsize$\,\pm0.02$} & $0.88${\scriptsize$\,\pm0.03$} \\
RWKV         & $0.12${\scriptsize$\,\pm0.02$} & $0.10${\scriptsize$\,\pm0.02$} & $0.08${\scriptsize$\,\pm0.02$} \\
RetNet       & $0.84${\scriptsize$\,\pm0.02$} & $0.70${\scriptsize$\,\pm0.03$} & $0.50${\scriptsize$\,\pm0.04$} \\
xLSTM        & $0.94${\scriptsize$\,\pm0.07$} & $0.88${\scriptsize$\,\pm0.11$} & $0.73${\scriptsize$\,\pm0.15$} \\
Naju (ours)  & $0.99${\scriptsize$\,\pm0.00$} & $0.98${\scriptsize$\,\pm0.02$} & $0.89${\scriptsize$\,\pm0.05$} \\
\bottomrule
\end{tabular}
\caption{Recency-only current-state tracking T4 (no final-update marker; scattered events; train @ $512$, length extrapolation), mean$\pm$std over $5$ seeds. Overwriting is required: Naju (ours, $0.89$), GLA ($0.88$), Mamba ($0.77$, high-variance), and xLSTM ($0.73$) are the strongest at $2048$; RWKV/HGRN/Transformer collapse. Together with T2, \textbf{Naju is strong on both} retention and overwriting; xLSTM comes closest but overwrites markedly less well ($0.73$). The ($150$ ep.)\ row is an extended $150$-epoch budget ($5$ seeds); as on T1/T2, it does not rescue T4 (train ${\approx}1.0$, test at chance). Indented sub-rows: final-epoch train / best-validation accuracy at the training length.}
\label{tab:sthard}
\end{table}
\paragraph{Recency-only current-state tracking (T4).}
T4 removes the explicit marker and repeatedly updates an entity, so the model
must infer which relevant update is the most recent and suppress stale values.
Events are scattered throughout the sequence, making the task jointly require
selective overwriting and subsequent retention of the updated value.

At length $2048$, Naju and GLA obtain comparable mean accuracies of $0.89$
and $0.88$, followed by Mamba at $0.77$ and xLSTM at $0.73$. Mamba also
shows substantial seed variability ($\pm0.27$). The ranking therefore differs
markedly from T2: GLA performs strongly on overwriting but poorly on retention,
whereas xLSTM performs strongly on retention but less strongly on overwriting.
Taken together, Naju provides the strongest observed balance, with a
worst-axis accuracy of
\begin{equation}
\min(\mathrm{T2},\mathrm{T4})
=
\min(0.99,0.89)
=
0.89 \notag
\end{equation}
at length $2048$. This joint comparison is summarized in
Section~\ref{sec:diag_summary} and Figure~\ref{fig:frontier}.

\paragraph{Residual difficulty at long overwrite distances.}
Bucketing the T4 examples by the distance from the answer-determining update
to the query shows that the most difficult regime is the \emph{far} bucket.
The model must suppress earlier values and then retain the latest value across
many intervening distractor events. This contrasts with T2, where bindings
are written once and never replaced. The remaining T4 error therefore
identifies headroom in combining precise overwriting with long-distance
retention, rather than in retention alone.

\section{Recurrent State Sizes of the Compared Models}
\label{app:state}

This appendix derives the \textbf{State} column of
Table~\ref{tab:summary}. We count the principal recurrent memory carried per
layer during sequential operation. This quantity excludes short convolution
buffers and small architecture-specific auxiliary states, and should not be
confused with the peak training memory in Table~\ref{tab:throughput}, which
is dominated by activations and backend-specific intermediates.

\paragraph{Formulas by architecture family.}
\begin{itemize}
    \item \textbf{Vector SSMs} (Mamba, Mamba-2, Naju): each of the
    $d_{\text{inner}}=E\,d_{\text{model}}$ channels carries a
    $d_{\text{state}}$-dimensional state,
    \[
    S
    =
    d_{\text{inner}}d_{\text{state}}
    =
    E\,d_{\text{model}}d_{\text{state}},
    \]
    with expansion $E=2$ throughout.

    \item \textbf{Matrix-valued recurrent memories}
    (GLA, RWKV, RetNet, xLSTM): each of $h$ heads carries a
    $(d_k/h)\times(d_v/h)$ matrix state,
    \[
    S
    =
    h\cdot\frac{d_k}{h}\cdot\frac{d_v}{h}
    =
    \frac{d_kd_v}{h},
    \]
    where $d_k=\kappa_kd_{\text{model}}$ and
    $d_v=\kappa_vd_{\text{model}}$ are the key and value widths used by
    the evaluated implementations.

    \item \textbf{Element-wise recurrences} (HGRN): one scalar recurrent
    state per channel,
    \[
    S=d_{\text{model}}.
    \]

    \item \textbf{Softmax attention} (Transformer): no fixed-size recurrent
    memory exists. During autoregressive inference, the KV cache stores the
    keys and values of all preceding tokens,
    \[
    S(L)=2L\,d_{\text{model}}
    \]
    per layer, and therefore grows as $\mathcal{O}(L)$. At $L=2048$ and
    $d_{\text{model}}=256$, this corresponds to $1{,}048{,}576$ floats,
    which is $64\times$ the principal recurrent state of Naju in the
    evaluated configuration.
\end{itemize}

\begin{table}[t]
\centering
\small
\begin{tabular}{llll}
\toprule
\textbf{Model} & \textbf{Configuration} &
\textbf{Computation} & \textbf{State/layer} \\
\midrule
Transformer
& $d_{\text{model}}=256$, $h=4$
& $2L\cdot256$
& $\mathcal{O}(L)$ \\
Mamba
& $d_{\text{model}}=256$, $d_{\text{state}}=16$
& $512\times16$
& $8{,}192$ \\
Mamba-2
& $d_{\text{model}}=128$, $d_{\text{state}}=64$
& $256\times64$
& $16{,}384$ \\
HGRN
& $d_{\text{model}}=128$, expand $1$
& $128\times1$
& $128$ \\
GLA
& $d_{\text{model}}=128$, $h=4$,
  $\kappa_k=0.5$, $\kappa_v=1$
& $4\times16\times32$
& $2{,}048$ \\
RWKV
& $d_{\text{model}}=128$, $h=4$,
  $\kappa_k=0.5$, $\kappa_v=1$
& $4\times16\times32$
& $2{,}048$ \\
RetNet
& $d_{\text{model}}=128$, $h=4$,
  $\kappa_k=1$, $\kappa_v=2$
& $4\times32\times64$
& $8{,}192$ \\
xLSTM
& $d_{\text{model}}=128$, $h=4$, proj.\ $2$
& $4\times64\times64$
& $16{,}384$ \\
Naju (ours)
& $d_{\text{model}}=128$, $d_{\text{state}}=64$
& $256\times64$
& $16{,}384$ \\
\bottomrule
\end{tabular}
\caption{Principal per-layer recurrent memory at the configurations used in
the diagnostic suite. The counts exclude short convolution buffers and small
architecture-specific auxiliary states. In particular, the evaluated mLSTM
also carries a vector normalizer and scalar stabilization state, which are
small relative to its matrix memory. The Transformer entry instead reports
the sequence-length-dependent KV cache.}
\label{tab:state_sizes}
\end{table}

\paragraph{State size alone does not explain the suite ranking.}
Mamba-2, xLSTM, and Naju each have a principal recurrent state of
$16{,}384$ values per layer in the evaluated configurations, but exhibit
substantially different T2 and T4 behavior. At length $2048$, Mamba-2
obtains $0.33$ on T2, xLSTM obtains $1.00$ on T2 and $0.73$ on T4, and
Naju obtains $0.99$ and $0.89$, respectively. Conversely, GLA uses a
smaller matrix state but achieves $0.88$ on T4. Thus, state size alone does
not determine the retention--overwrite ranking; the state-update and gating
mechanisms also play an important role. State size can still affect
associative-recall capacity, so capacity-oriented benchmarks such as MQAR
complement the mechanism-oriented diagnostic suite.

\paragraph{State memory versus training memory.}
The figures above describe the principal memory carried during sequential
inference. Peak training memory in Table~\ref{tab:throughput} is instead
determined largely by activations and backend-specific parallel
intermediates. The evaluated parallel xLSTM implementation materializes
$L\times L$ decay and query--key intermediates and runs out of memory at
$L=8192$, despite having a principal recurrent-state size comparable to
Naju's. This is an implementation-specific training-memory result rather
than a limitation implied by the recurrent mLSTM state itself. In contrast,
the evaluated scan-based SSM implementations avoid materializing an
$L\times L$ intermediate and exhibit linear memory scaling with sequence
length.

\section{WikiText-103 Learning-Rate Selection}
\label{app:wt103_lr}

We select learning rates using a $250$M-token pilot run with seed $1$ under
the protocol of Section~\ref{sec:wt103}. The primary candidate grid is
\begin{equation}
\left\{
10^{-3},
2{\times}10^{-3},
3{\times}10^{-3},
4{\times}10^{-3}
\right\}. \notag
\end{equation}
Additional higher rates are evaluated for the recurrent models when the pilot
curve is still improving near the upper end of the primary grid. Blank entries
in Table~\ref{tab:wt103_lr} indicate learning rates that were not evaluated.
Model selection uses validation perplexity only; test perplexity is never used
to select a learning rate.

\begin{table*}[t]
\centering
\small
\setlength{\tabcolsep}{3.5pt}
\begin{tabular}{lcccccccc}
\toprule
\textbf{Model}
& $10^{-3}$
& $2{\times}10^{-3}$
& $3{\times}10^{-3}$
& $4{\times}10^{-3}$
& $5{\times}10^{-3}$
& $6{\times}10^{-3}$
& $7{\times}10^{-3}$
& $8{\times}10^{-3}$ \\
\midrule
Transformer
& $33.59$
& $\underline{\mathbf{31.27}}$
& $31.37$
& --
& --
& --
& --
& -- \\
Mamba ($N\,16$)
& $38.41$
& $34.02$
& $32.44$
& $\underline{\mathbf{31.76}}$
& $32.88$
& --
& --
& -- \\
Mamba ($N\,64$)
& --
& --
& $33.24$
& $\mathbf{32.31}$
& $31.90$
& $31.47$
& $31.36$
& $\underline{31.27}$ \\
Mamba-2 ($N\,64$)
& $36.34$
& $32.92$
& $31.97$
& $\underline{\mathbf{31.50}}$
& $33.19$
& --
& --
& -- \\
Naju ($N\,64$)
& $35.84$
& $32.36$
& $31.10$
& $\underline{\mathbf{30.83}}$
& $31.56$
& --
& --
& -- \\
\bottomrule
\end{tabular}
\caption{\textbf{WikiText-103 learning-rate pilot.}
Best validation perplexity after $250$M training tokens on the full validation
set. Underline marks the lowest pilot perplexity among the rates evaluated in
each row, and bold marks the learning rate used in the final $1.2$B-token
runs. Dashes denote rates that were not evaluated. Each model uses its
lowest-validation pilot rate except Mamba with $N=64$. For this state-size
comparison, we retain the $4{\times}10^{-3}$ rate selected for Mamba with
$N=16$, so that increasing $N$ is not accompanied by a simultaneous change
in the optimization configuration.}
\label{tab:wt103_lr}
\end{table*}

\section{The Native CUDA Scan}
\label{app:cuda_scan}
This appendix describes three Naju scan implementations: a Triton reference
backend, a Blackwell-optimized sequential CUDA backend
(\texttt{blackwell\_seq}) used for the forward-latency measurements in
Table~\ref{tab:throughput}, and a fused chunk-parallel backend used for the
training-efficiency measurements in Table~\ref{tab:wt103_eff}. The sequential
CUDA implementation also serves as a numerical reference and fallback for the
chunk-parallel backend. 
The three backends implement the same diagonal affine scan core,
\begin{equation}
x_n
=
f_n\odot x_{n-1}
+
i_n\odot
\left(
h_n B_n^{\top}
\right).
\notag
\end{equation}
The corresponding recurrent-memory readout is
$
y_n^{\mathrm{mem}}
=
\alpha x_n C_n$,
while the feedthrough term $D\odot h_n$ and the surrounding output
modulation and projection are applied consistently across backends.

\paragraph{Triton reference backend.}
The Triton implementation evaluates the recurrence sequentially and stores the
state trajectory in HBM for use during backward propagation. Its simple
implementation serves as the numerical reference, but the materialized state
history incurs substantial memory traffic.

\paragraph{Blackwell-optimized sequential CUDA backend.}
The \texttt{blackwell\_seq} backend evaluates the recurrence one position at a
time in a fused CUDA kernel. The running state is kept in registers, and the
backward pass reconstructs intermediate states from sparse checkpoints rather
than storing the complete trajectory. This backend is optimized for
forward-only execution and is used in Table~\ref{tab:throughput}.

\paragraph{Fused chunk-parallel backend.}
The chunk-parallel backend partitions the sequence into chunks of length
$Q=64$. Within each chunk, affine state propagation and output computation are
evaluated using parallel matrix operations; only the carry between chunk
boundaries remains sequential. The backward pass applies the corresponding
reverse-time affine recurrence analytically. This backend is used for the
training-efficiency measurements in Table~\ref{tab:wt103_eff}.

All three implementations preserve the same Naju recurrence and agree with the
reference outputs and gradients within the evaluated floating-point tolerance.

\end{document}